\documentclass[11pt,a4paper]{article}
\usepackage[hyperref]{emnlp2020}
\usepackage{times}
\usepackage{latexsym}

\usepackage{algorithm}
\usepackage{pgfplots}
\usepackage{subfig}
\usepackage{url}
\usepackage{booktabs}
\usepackage{multirow}
\usepackage[noend]{algpseudocode}
\usepackage{amsmath,amssymb,amsfonts}
\usepackage{bbm}
\usepackage{balance}

\usepackage{microtype}

\aclfinalcopy 
\def\aclpaperid{1595} 

\providecommand{\ie}{\emph{i.e.,} }
\providecommand{\eg}{\emph{e.g.,} }
\providecommand{\parab}[1]{\paragraph{#1}}

\newcommand{\minus}{\scalebox{1.15}[1.0]{$-$}}

\DeclareMathOperator*{\argmin}{\argmin}

\newcommand*{\affaddr}[1]{#1} 
\newcommand*{\affmark}[1][*]{\textsuperscript{#1}}
\newcommand*{\email}[1]{\texttt{#1}}

\newcommand{\karthik}[1]{{\color{magenta} [Karthik: {#1}]}}

\title{Robust and Interpretable Grounding of Spatial References \\ with Relation Networks}

\author{%
Tsung-Yen Yang\affmark[1], Andrew S. Lan\affmark[2],  Karthik Narasimhan\affmark[1]\\
\affaddr{\affmark[1]Princeton University, Princeton, NJ}\\
\affaddr{\affmark[2] University of Massachusetts, Amherst, MA}\\
\email{\{ty3, karthikn\}@princeton.edu}, 
\email{andrewlan@cs.umass.edu}
}

\date{}

\begin{document}
\maketitle
\begin{abstract}
Learning representations of spatial references in natural language is a key challenge in tasks like autonomous navigation and robotic manipulation.
Recent work has investigated various neural architectures for learning multi-modal representations for spatial concepts.
However, the lack of explicit reasoning over entities makes such approaches vulnerable to noise in input text or state observations.
In this paper, we develop effective models for understanding spatial references in text that are robust and interpretable, without sacrificing performance.
We design a text-conditioned \textit{relation network} whose parameters are dynamically computed with a cross-modal attention module to capture fine-grained spatial relations between entities. 
This design choice provides interpretability of learned intermediate outputs.
Experiments across three tasks demonstrate that our model achieves superior performance, with a 17\% improvement in predicting goal locations and a 15\% improvement in robustness compared to state-of-the-art systems.
%
\footnote{Code is available at \url{https://sites.google.com/view/robust-relation-net/home}.
}
\end{abstract}

\section{Introduction}
\label{sec:introduction}

\begin{figure}[!h]
\centering
\includegraphics[width=\linewidth]{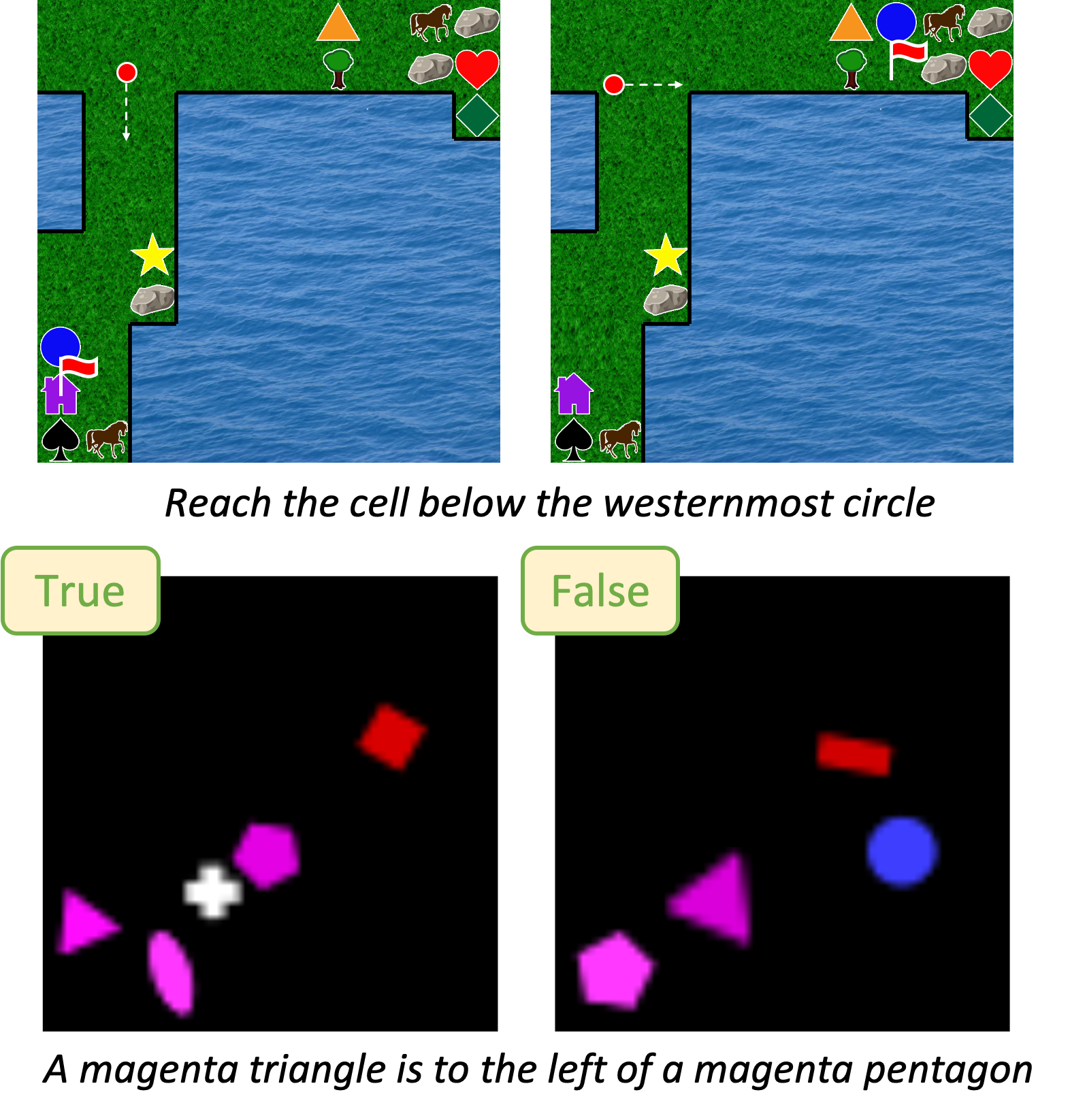}
\vspace{-0.3in}
\caption{Two different tasks requiring joint spatial reasoning over observation and text -- \textbf{(top)} the same instruction may specify different goal locations, depending on the map (red flags = goals); \textbf{(bottom)} the same statement may be true or false depending on the image. 
}
\label{fig:example}
\end{figure}

Grounding spatial references in text is essential for effective human-machine communication through natural language. Spatial reasoning is ubiquitous in many scenarios such as autonomous navigation~\cite{macmahon2006walk,vogel2010learning}, situated dialog~\cite{skubic2002using} and robotic manipulation~\cite{landsiedel2017review}. Despite tremendous applicability, understanding spatial references is a highly challenging task for current natural language processing (NLP) systems, requiring a solid contextual understanding of language dependent on other observations from the environment. Figure~\ref{fig:example} demonstrates two tasks where the interpretation of the instruction or statement changes completely with the observation provided. In the first example, the \emph{westernmost circle} may lie to the left or right of the navigating agent's starting location (red dots). In the second, the validity of the statement depends on the relative orientation of the relevant objects -- the \textit{triangle} and the \textit{pentagon}.



Some of the earliest work in this field~\cite{herskovits1987language,regier1996human} investigated the grounding of spatial prepositions (\eg ``above'', ``below'') to perceptual processes like visual signals. Such early grounding efforts were limited by computational bottlenecks but several deep neural architectures have been recently proposed that jointly process text and visual input~\cite{janner2017representation,misra2017mapping,bisk2016natural,liu2019representation,jain2019stay,gaddy2019pre,hristov2019disentangled,yu2018interactive}. While these approaches have made significant advances in improving the ability of agents at following spatial instructions, they are either not easily interpretable or require pre-specified parameterization to induce interpretable modules~\cite{bisk2018learning}. Moreover, their end-to-end formulations and lack of explicit entity reasoning make them susceptible to noise and perturbations in the input data, as we demonstrate in our experiments (Section~\ref{sec:results}).

In this paper, we develop a model to perform robust and interpretable grounding of spatial references in text. 
Our objective is to understand and analyze how to ground spatial concepts in a robust and interpretable way.
In particular, we focus on the class of \emph{deictic spatial references}~\cite{logan1996computational}, which specify a location or object in terms of one or more \emph{reference objects}. Our key idea is to decompose the spatial reasoning process into two important steps: (1) identifying the reference object(s) (\eg \textit{rock, circle}) from the instructions, and (2) accurately inferring the spatial direction (\eg \textit{left, top-right}) of the goal with reference to that object. 

We use a relation network~\cite{santoro2017simple} to implicitly enable this factorization by computing representations for each location in the environment based on its interactions with neighboring entities. 
The parameters of the relation network are dynamically derived from a vector representation of the input text followed by an attention module conditioned on the observations of the environment.
This architecture provides three key benefits:
(1) the dynamically computed parameters of the relation network enable fine-grained modeling of spatial references, (2) since the model considers relations between pairs of entities, it is more robust to noisy inputs, and (3) the explicit multi-modal representations learned for each entity pair are highly interpretable.
%
%
%
%
%
%
%
%
%
%

%
We empirically test our model on three different task settings -- classification, value map regression and reinforcement learning (RL) for navigation, and compare its performance to existing state-of-the-art methods. We find that our approach is competitive with or outperforms the baselines under several different evaluation metrics. For example, in the navigation task with RL, our model obtains up to 13.5\% relative improvement in policy quality over the best performing baseline. Our approach is also more robust to noisy inputs -- for instance, after adding unseen objects as noise to a value map regression task, our model's performance degrades by only around 10\% compared to over 20\% for the best baseline. Finally, we also present several visualizations of relation and value maps produced by the model, which demonstrate appropriate grounding of reference objects as well as spatial words. 

\section{Related Work}
\label{sec:related}

The role of language in spatial reasoning has been explored since the 1980s~\cite{herskovits1987language,logan1996computational,regier1996human,regier2001grounding}. Most early papers dealt with the question of representing spatial prepositions~\cite{herskovits1987language,coventry2004spatial,coventry2004saying} and grounding them to spatial templates~\cite{logan1996computational}. \citet{regier2001grounding} introduced the influential attention vector-sum model which accurately predicted human spatial judgements for words like ``above'' and ``below''. The use of neural networks to computationally ground spatial terms to geometric orientations was first explored by \citet{regier1996human} and later by \citet{cangelosi2005grounding}. While spatial reasoning in general is a wide-ranging problem, in this paper, we focus on grounding \textit{deictic spatial references} in third person, which involve referring to a goal location using one or more referent objects~\cite{logan1996computational}.

\parab{Spatial Reasoning in Text.}
Reasoning about spatial references has been explored in various contexts such as instruction following for 2-D and 3-D navigation~\cite{macmahon2006walk,vogel2010learning,chen2011learning,artzi2013weakly,kim2013adapting,Andreas15Instructions,fried2018speaker, liu2019representation,jain2019stay,gaddy2019pre,hristov2019disentangled,chen2019touchdown} and situated dialog for robotic manipulation~\cite{skubic2002using,kruijff2007situated,kelleher2009applying,landsiedel2017review}. Most of these approaches utilize supervised data, either in the form of policy demonstrations or target geometric representations. 

More recent work has demonstrated the use of RL in navigation tasks that require spatial reasoning for goal prediction. \citet{misra2017mapping} use a factored approach to process both text and visual observations in parallel, before fusing the representations to capture correspondences. \citet{janner2017representation} use a recurrent network to generate vector representations for the text, which then serve as parameters for an observation processing module (\eg convolutional neural network (CNNs)).  A similar architecture called LingUNet was employed to process observation frames in 3-D navigation tasks~\cite{misra2018mapping,blukis2018mapping}, producing probability distributions over goals that are used by the agent to predict action sequences. These pieces of work can be viewed as using forms of feature-wise transformations~\cite{dumoulin2018feature-wise} in neural architectures. While we also employ a similar form of text-conditioning, our model processes observations using a relation network~\cite{santoro2017simple} (instead of convolutions) which allows us to capture spatial references in a fine-grained manner, robust to noise. 

\parab{Interpretable Spatial Reasoning.}
\citet{ramalho2018encoding} learn viewpoint-invariant representations for spatial relations expressed in natural language. They propose a multi-modal objective to generate images of scenes from text descriptions, and show that the learned representations generalize well to varying viewpoints (and their corresponding descriptions).
\citet{bisk2018learning} learn interpretable spatial operators for manipulating blocks in a 3-D world. They build a model to explicitly predict the manipulation operators and objects to manipulate, which are used as inputs to another neural network that predicts the final location of each object. The manipulation operators can then be associated with canonical spatial descriptions (\eg below, south). We focus on demonstrating learned associations between the text representations and visual observations instead of the manipulation operators (actions). Moreover, we also consider the RL setting while both papers above require full supervision.

\section{Framework and Design}
\label{sec:reinforcementLearning}

\begin{figure*}[t]
\centering
\includegraphics[width=1.0\linewidth]{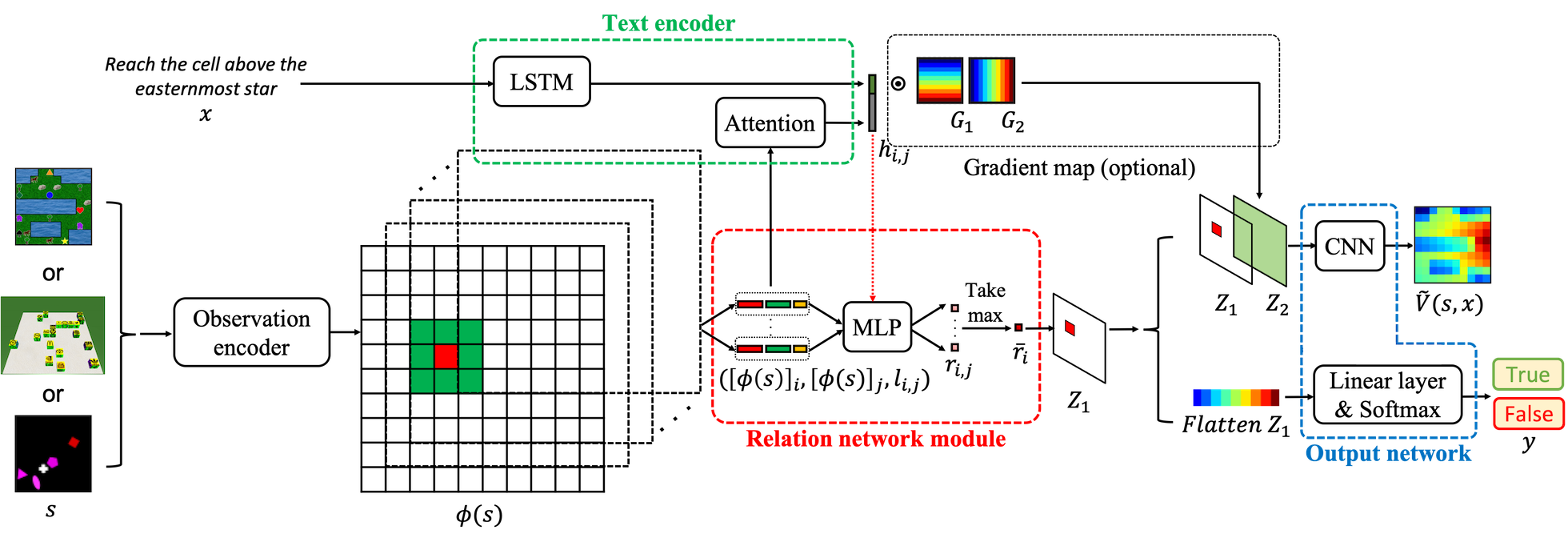}
\vspace{-0.2in}
\caption{\textbf{(Model overview)} Our architecture consists of three parts -- (1) a relation network takes a tensor $\phi(s)$ provided by the observation encoder as input and produces a relation map $Z_1$; (2) a text encoder (LSTM+attention) converts the text into a vector $h_{i, j}$ for each pair of cells $\langle s_i, s_j\rangle$ in the observation. This $h_{i,j}$ is split and reshaped into the parameters of the relation network module; (3) an output network that takes $Z_1$ as input and produces the final outputs to be predicted, depending on the task.
The max operator in the relation network allows the model to attend to the pair that has a highest value. 
This enables robustness to the noise and interpretability of the model. 
}
\label{fig:ourModel}
\end{figure*} 

\subsection{Setup}
We consider 2-D map-like fully observable environments, where the accompanying text contains spatial references that are key to understanding the goal location. This text contains references to objects or landmarks in the world, as well as relative spatial positions such as \textit{``above"} or \textit{``to the bottom left of"}. We do not assume access to any ontology of entities or spatial references -- the agent has to learn representations using feedback from the task. We consider two settings  -- (1) supervised learning and (2) reinforcement learning.

\parab{Supervised Learning.}
In the supervised scenario, we assume access to data with ground-truth annotation for the quantity to be predicted. This can be either (1) classification labels (\eg as in ShapeWorld~\cite{andreas2017learning}), or (2) value maps (\eg as in PuddleWorld~\cite{janner2017representation}). In this case, the model takes the inputs of an observation map $s \in S$ and a text instruction $x \in X$ and predicts the required outputs. 

\parab{Reinforcement Learning.}
We also consider an instruction-following scenario where the main source of supervision is a scalar reward provided upon successful completion of the task. We employ a standard Markov decision process (MDP) framework $<S, A, X, T, R>$, where $S$ is the set of states, $A$ is the set of the actions, $X$ is the set of possible text instructions, $T$ is the transition probability of the environment and $R$ is the reward function. Given a text instruction $x \in X$ and the current state $s$, the agent takes actions according to a policy  $\pi(a|s,x): S \times X \rightarrow A$, which transitions the environment to the next state $s'$ according to the state transition model $T(s'|s,a,x)$. 
For simplicity, we assume $T$ is deterministic.
This RL setup is inherently harder than supervised learning due to the sparse and weak feedback.

\subsection{Model}
\label{sec:model}
Any model that can ground spatial references must have the ability to learn flexible, compositional representations for text and effectively fuse it with visual observations. Prior work has explored neural architectures with feature-wise conditioning, where the text representations are used to dynamically induce parameters of Convolutional Neural Networks (CNNs) that process the observations~\cite{janner2017representation,misra2018mapping}. While CNNs are useful to capture spatial invariances, they do not provide fine-grained reasoning abilities since the convolution operates at the coarse level. 

To this end, we use a text-conditioned Relation Network to compute representations over the observations. Unlike the relation network in~\citet{santoro2017simple}, the parameters of our relation network are dynamically initialized using the text instruction provided, allowing us to implicitly factorize the reasoning process into locating the reference object(s), and inferring the goal location relative to the reference object(s). Our architecture consists of three main components -- \textit{(a) a text encoder $\psi$, (b) a text-conditioned relation network (RNet) and (c) a task-dependent output network $\tau$.}  Figure~\ref{fig:ourModel} provides an overview of the entire architecture. For ease of exposition, we will first describe the RNet, followed by the other modules. 

\parab{(a) Relation Network (RNet).}
Assume the input observation $s$ to be a 2-D matrix.\footnote{Though similar architectures can be built for 3-D inputs, we focus on 2-D observations in this paper.} First, we convert this matrix into a 3-D tensor $\phi(s)$ by encoding each cell as a vector through an observation encoder (similar to a word embedding).
If the element in $s$ is the index of the entity, the observation encoder is an embedding network.
On the other hand, if $s$ is a raw image pixels, the observation encoder is a CNN.
Next, we feed this tensor into our relation network module $f$, which computes representations for each cell in the 2-D grid as a function of its neighboring cells. This is done using a multilayer perceptron (MLP) that computes a scalar \textit{relation score} $r_{i, j}$ for each pair of neighboring cells $s_i, s_j$ as:
\[
r_{i,j} = f([\phi(s_i)], [\phi(s_j)],l_{i,j})
\]
where $[\phi(s_i)] \in \mathbb{R}^{k}$ is the embedding for cell $s_i$, and $l_{i,j}\in \mathbb{R}$ is the encoding for the relative location of these two cells.
Intuitively, this relation score represents the relevance of the pair of objects (and their positions) in locating the goal. 
We then perform max operator over all the $r$-scores associated with each cell to build a relation map $Z_1 \in \mathbb{R}^{m\times n}$, each cell of which is computed as:
\[
[Z_{1}]_{i} = \max\limits_{j\in \mathcal{N}(i)} r_{i,j},
\]
where $m$ and $n$ are the size of the input observation, and $\mathcal{N}(i)$ is the set of neighbors of cell~$i.$\footnote{Depending on the tasks, our approach can be extended to handle longer-range relations by considering cells that are more than 2 cells away (though this may require some form of refinement (\eg beam search) to choose appropriate entities).
}
The max operator allows the model to attend to the pair that has the highest relation score.
Finally, since the processing of the observation should depend on the instruction, we dynamically predict all the parameters of the RNet (\ie $f$) for each input $s_i, s_j$ using the text provided (details below).

\parab{(b) Text Encoder.}
We use an LSTM recurrent network~\cite{hochreiter1997long} to convert the instruction text $x$ into a vector $h$, which is taken to be the weighted combination of the output state of the LSTM after processing the entire instruction. This vector $h$ is used to dynamically initialize all the parameters of the relation network module. We simply choose the size of $h$ to be equal to the total number of parameters in the RNet and split and reshape $h$ accordingly. 
For example, if RNet $f$ is a two layers of a MLP with the size of $(a_1,a_2),$ the size of vector $h$ is $(2k+1)\cdot a_1 + a_1\cdot a_2 + a_2\cdot 1.$ 
We then take the first $(2k+1)\cdot a_1$ components of $h$ and reshape it into a 2-D matrix which is the weight of $f$ in the first layer, and so on.
As a result, the computation of the first layer of $f$ is $W_1[\phi(s_i);\phi(s_j);l_{i,j}]$ followed by an activation function, where $W_1\in\mathbb{R}^{a_1\times (2k+1)}.$
More examples are in Appendix~\ref{section:implmentation}.
Therefore, for each different instruction, RNet will process the observations using a different set of parameters. 

\parab{Attention:} To encourage tighter coupling between the processing of text and observations, we also add an attention module. We compute the text representation $h$ as a weighted combination of each word's representation, where the weights are influenced by the current pair of cells being considered by RNet. Specifically, when processing cells $s_i, s_j$, we compute
$h_{i,j}$ as: 
\[
h_{i,j}=\sum_{k=1}^{L}\alpha_{i,j}(k)h(k),
\]
where $\alpha_{i,j}(k)\propto \mathrm{e}^{h(k)^T([\phi(s_i)], [\phi(s_j)])}$ and $L$ is the length of the instruction text. Note that this means the MLP parameters (\ie the relation network module $f$) will depend not only on the instruction, but also on the pair of cells $s_i, s_j.$ 


\parab{(c) Output Network.}
The final component of our architecture is a task-dependent output network $\tau$, whose form varies according to the task and the type of supervision. For the tasks we consider, we develop two variants of this:\\
(1) For classification tasks, we simply flatten $Z_1$ into a vector and pass it through a linear layer followed by a Softmax function to predict the class. \\
(2) For predicting value maps (in both supervised and reinforcement learning), we use convolution operations. Following~\citet{janner2017representation}, we add two global gradient maps (horizontal and vertical) which have been shown to help global spatial references (\eg ``\textit{the easternmost house}''). We use the last three elements of $h$ to produce the coefficients $\beta_1, \beta_2, \beta_3$ and produce a map $Z_2$:
\[ Z_2 = \beta_1 \cdot G_1 + \beta_2 \cdot G_2 + \beta_3 \cdot J, \]
where $J\in\mathbb{R}^{m\times n}$ is an all-ones matrix and $\beta$s are predicted using the text encoder as described above. Then, we concatenate $Z_{1}$ (the output of RNet) with $Z_2$ and feed this tensor ($[Z_{1}; Z_{2}]\in \mathbb{R}^{m\times n\times 2}$) into a convolutional layer to predict the value function $\tilde{V}(s,x)$.
We call our model \textit{t-RNetAttn}.


\subsection{Learning}
For all cases below, we train all the parameters of our model jointly, including those in the text encoder, RNet and output networks. 

\parab{Supervised Learning.}
As previously mentioned, we consider two supervised learning scenarios -- classification and value map prediction (a regression task). 
For classification, we train our model using softmax loss:
\begin{equation}
\begin{split}
\mathcal{L}_1(\Theta) =-\text{E}_{s,x\sim\mathcal{D}}[y\log(p)], \nonumber
\end{split}
\end{equation}
where $y$ is the ground truth label, and $p$ is the predicted probability of certain class.

For value maps prediction, we minimize the mean squared error (MSE) between the model's prediction and the ground truth:
\begin{equation}
\begin{split}
\mathcal{L}_2(\Theta) =\text{E}_{s,x\sim\mathcal{D}}\bigg[ \big(\tilde{V}_{\Theta}(s ,x) - {V}(s ,x)\big)^2 \bigg], \nonumber
\end{split}
\end{equation}
where $\Theta$ denotes the parameters in the entire model, and $V(s, x)$ is the ground truth.

\parab{Reinforcement Learning.} 
In the RL scenario, we explore the environment using the predicted value map. With the collected trajectories, we then perform fitted Value iteration~\cite{munos2008finite}:
\begin{equation}
\begin{split}
\mathcal{L}_3(\Theta) =\text{E}_{(s,a,r,s')\sim\mathcal{\tau}}&\bigg[ \tilde{V}_{\Theta}(s ,x) \\ 
- \bigg(r +  \gamma \max\limits_a & E_{s' \sim T(s'|s,a)}  \tilde{V}_{\Theta'}(s',x)\bigg) \bigg], \nonumber
\end{split}
\end{equation}
where $\Theta$ denotes the parameters of the entire model, and $\Theta'$ denotes a set of target parameters that are periodically synced with $\Theta$. 

\section{Experimental Setup}
\label{sec:exp}

\begin{table*}[t!]
\centering
\centering
\subfloat[Classification]{
\scalebox{0.8}{
\begin{tabular}{l*{1}{c}r}
 &\multicolumn{1}{c}{SW}\\ \cmidrule(lr){2-2} & ACC $\uparrow$\\ \hline
t-VGG~\cite{andreas2017learning}  & 0.71 \\ 
t-RNetAttn (ours)  & \bf{0.72}\\  
\bottomrule
\end{tabular}
}}
\subfloat[Goal navigation with RL]{
\scalebox{0.8}{\raisebox{+1.37ex}{
\begin{tabular}{l*{6}{c}r} 
&\multicolumn{2}{c}{PW local} &\multicolumn{2}{c}{PW global} &\multicolumn{2}{c}{ISI }   
\\ \cmidrule(lr){2-3} \cmidrule(lr){4-5}\cmidrule(lr){6-7}
& PQ $\uparrow$ & MD $\downarrow$  & PQ $\uparrow$ & MD $\downarrow$ & PQ $\uparrow$ & MD $\downarrow$ \\ \hline
t-CNN~\cite{janner2017representation}                             & 0.89 & \bf{2.03}  & 0.91 & \bf{3.80} & 0.74       & 3.94         \\
t-UVFA~\cite{schaul2015universal}                       & 0.56       & 4.71        & 0.62 & 6.28  & 0.15       & 4.61          \\ 

t-RNetAttn (ours)                             & \bf{0.91} & 2.10  & \bf{0.93} & 4.23 & \bf{0.84} & \bf{3.79} \\

\bottomrule
\end{tabular}}}
}

\subfloat[Value map regression]{
\scalebox{0.8}{
\begin{tabular}{l*{8}{c}r}
 &\multicolumn{3}{c}{PW local}
                                    &\multicolumn{3}{c}{PW global}  
                                    &\multicolumn{2}{c}{ISI}  \\ \cmidrule(lr){2-4} \cmidrule(lr){5-7}
                                    \cmidrule(lr){8-9}
                       & MSE  $\downarrow$ & PQ $\uparrow$ & MD $\downarrow$  & MSE  $\downarrow$ & PQ $\uparrow$ & MD $\downarrow$ & MSE  $\downarrow$ & MD $\downarrow$ \\ \hline
                                                            
                                                            t-CNN~\cite{janner2017representation}                         & 0.25 & \bf{0.94} & 2.34  & 0.41 & 0.89 & \bf{3.81}  & \bf{0.15} & \bf{3.14} \\
t-UVFA~\cite{schaul2015universal}                           & 3.23       & 0.57 & 4.97        & 1.90 & 0.62 & 5.31 & \minus  & 4.61  \\

t-RNetAttn (ours)                             & \bf{0.22} & \bf{0.94} & \bf{1.95}  &\bf{0.40}& \bf{0.91} & 3.82 & \bf{0.15}  & 3.43 \\

\bottomrule
\end{tabular}
}

}

\caption{Performance of all models on all three tasks -- classification, value map regression and goal navigation with RL (PW: PuddleWorld, SW: ShapeWorld, PQ: policy quality, MD: Manhattan distance, MSE: mean squared error)). Arrows denote higher or lower scores being better. Best values are in bold. 
}
\label{table:results}
\end{table*}

\parab{Tasks.}
We perform several empirical studies and compare our model with prior work in terms of accuracy and robustness. As previously mentioned, we focus on deictic spatial references, which involve talking about a location or object in terms of other referent objects. We consider three different prediction tasks with accompanying text. These tasks all involve joint reasoning over both observations and the text in order for a system to perform well. The tasks are as follows:\vspace{0.0in} \\ 
1. \textit{Classification}: Given an image and a text statement containing spatial references, predict whether the statement is applicable. We use data from ShapeWorld~\cite{andreas2017learning}, which contains images of abstract objects in various shapes and sizes, each paired with a statement about their relative positions and a True/False annotation. \vspace{0.0in} \\ 
2. \textit{Value map regression}: In this task, the input is a top-down view of a navigation environment along with a text instruction describing a goal location. The aim is to produce the optimal value function map with a value $V(s)$ for each location in the map with respect to the goal. For this task, we use two recently proposed instruction following datasets -- ISI~\cite{bisk2016natural} and PuddleWorld~\cite{janner2017representation}. ISI contains a set of blocks, each with a different company logo or number, with the instruction being to move a particular block to a desired location. PuddleWorld (PW) is a navigation environment with a positive reward for reaching the goal and negative rewards for stepping in puddles. PW consists of local instructions with local neighborhoods in references \eg \textit{``two cells to the left of the triangle''} and global instructions with description about the entire map \eg \textit{``the westernmost rock.''} The ground truth value maps for each instance are obtained using the value iteration algorithm~\cite{sutton1998introduction}.\vspace{0.0in} \\ 
3. \textit{Goal navigation with RL}: This is a variant of the previous task where the agent is not provided with ground truth value maps, and instead has to explore the environment, receive rewards (both positive and negative) and learn a policy to navigate to the goal conditioned on the text instruction. We make use of the same datasets as above, sans the value maps. 

These three tasks cover different application settings for spatial reasoning. While the prediction objective is different in each one, the input observations are also varied -- ShapeWorld uses pixels, while ISI and PuddleWorld are grid worlds. All three are fully observable 2-D environments, and do not contain first-person points of view or certain kinds of spatial references such as intrinsic relations~\cite{logan1996computational}. However, these environments are sufficient to demonstrate the accuracy, robustness and interpretability of our approach and there is nothing inherently preventing the model from generalizing to more scenarios (\eg 3-D, different points of view, etc.). More statistics on the datasets including train-test splits are provided in Table~\ref{table:dataset} in the appendix. 

\begin{figure*}[t]
\centering
\vspace{0.0in}
\includegraphics[width=0.95\linewidth]{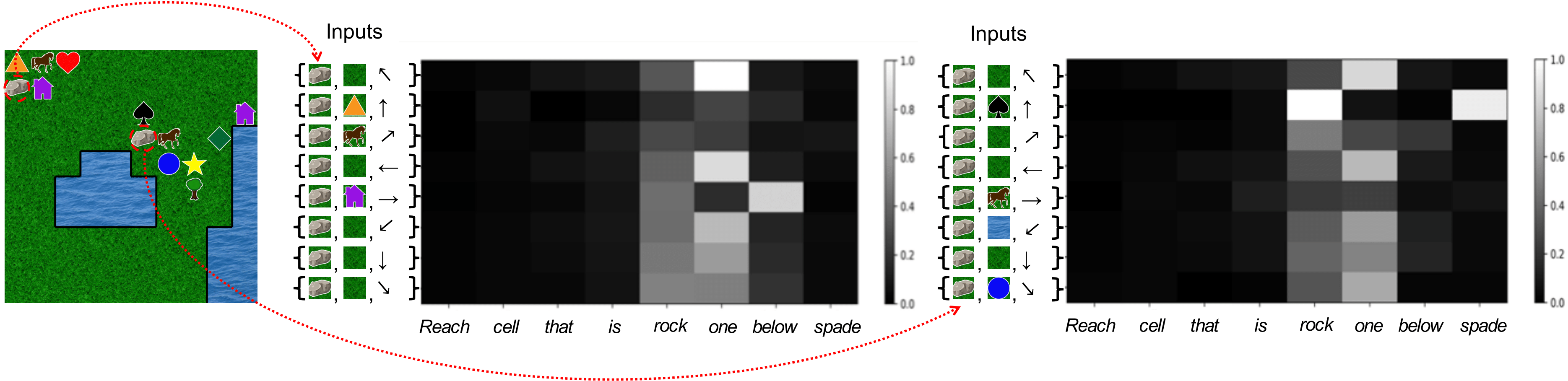}
\vspace{-0.1in}
\caption{Attention weights in the text encoder for different cell pairs in the observation and the instruction ``\textit{Reach cell that is rock one below spade}''. The attention weights on the left are for the rock in the upper left corner, and the ones on the right are for the rock in the middle. We see that the model attends correctly to \textit{rock} and \textit{spade} on the right, helping it locate the correct goal. (Note that in the attention weights on the right, the values of weights on the words ``\textit{rock}'' and ``\textit{spade}'' are large simultaneously when the spatial relation is mentioned in the text.)}
\label{fig:localEmbedding}
\end{figure*}

\begin{figure*}[t]
\centering
\vspace{-0.0in}
\subfloat[Observational noise]{\includegraphics[width=3.0in]{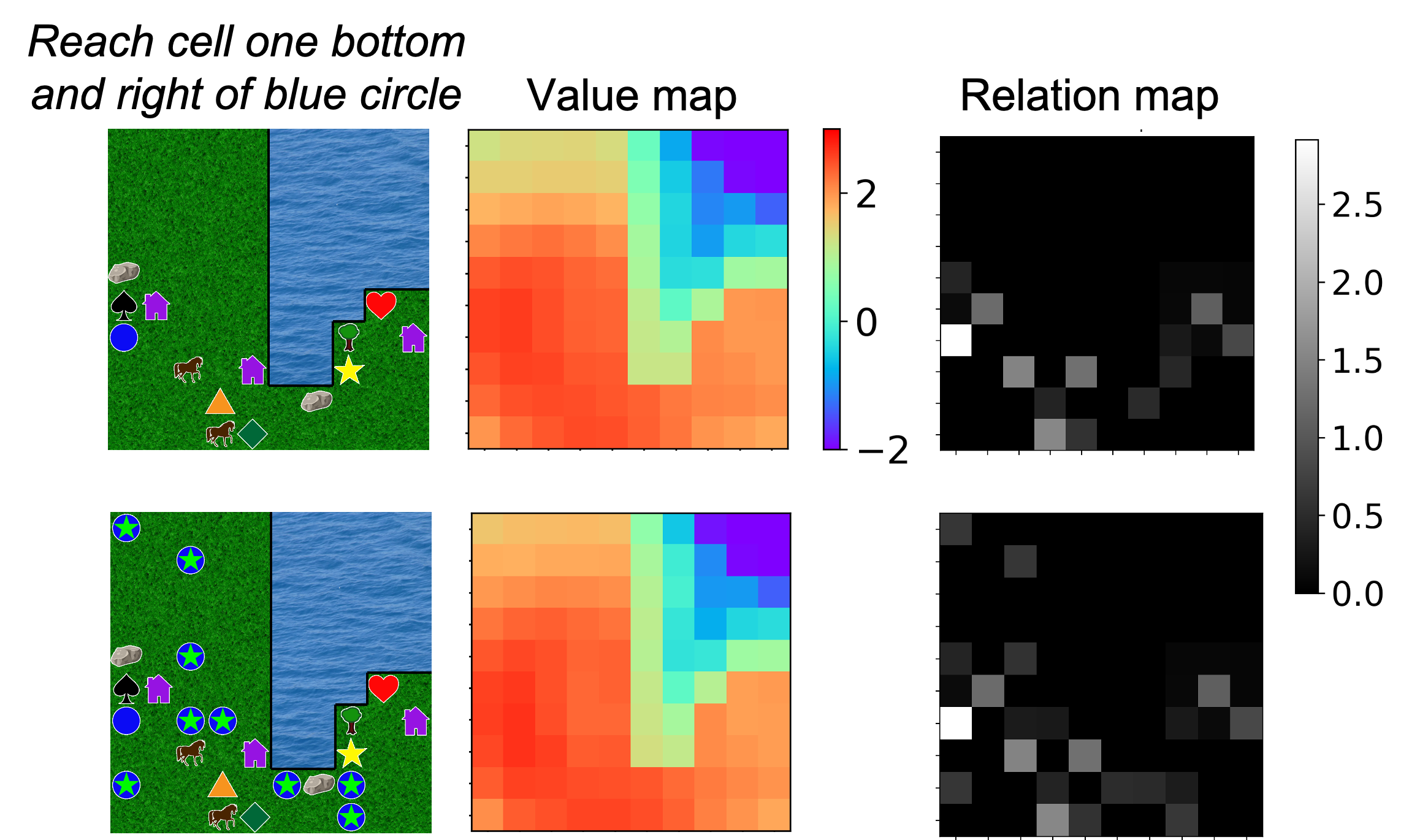}}
\subfloat[Textual noise]{\includegraphics[width=3.3in]{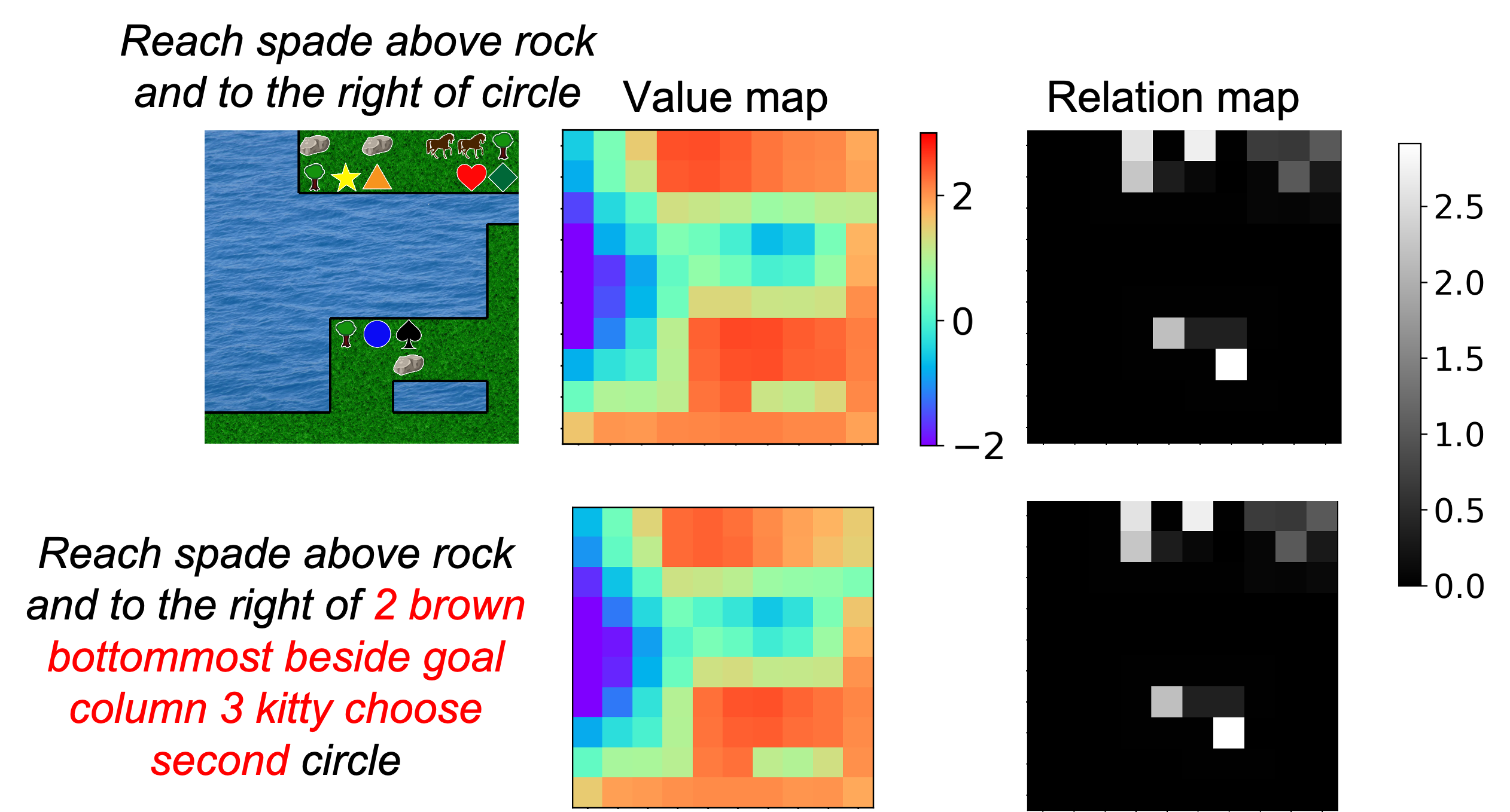}}
\vspace{-0.1in}
\caption{Visualization of value maps and relation maps after taking absolute values $|Z_1|$ from t-RNetAttn, without (top) and with observation and textual noise (bottom) in the PuddleWorld environment. Blue stars with circle are unseen objects which are not presented during training. Our approach produces sharper (magnitude-wise) $|Z_1|$ values for goal location and referent objects, and is almost undisturbed by noise.
}
\label{fig:noiseCaseEnv_inter}
\end{figure*}



\parab{Evaluation Metrics.}
We use several different quantitative metrics to evaluate the models:\vspace{0.0in} \\ 
1. \textit{Accuracy} (ACC) of predictions for the binary classification task.\vspace{0.0in} \\ 
2. \textit{Mean square error} (MSE) between the predicted and ground truth value maps for the regression task.\vspace{0.0in} \\ 
3. \textit{Policy quality} (PQ), which is a normalized reward score obtained by the agent's policy compared to the optimal policy~\cite{schaul2015universal}.\vspace{0.0in} \\ 
4. \textit{Manhattan distance} (MD) which measures the distance between the agent's final position and the ground truth goal location. 

The last two measures (PQ and MD) are naturally applicable to the navigation task with RL, but we also apply them to the regression task by inducing a policy from the predicted value map as \[
\pi(s) = \arg\max_a R(s, a) + \gamma T(s' | s, a) \tilde{V}(s').
\]

\parab{Baselines.}
For binary classification on  ShapeWorld, we use the text-VGG net (t-VGG) from \citet{andreas2017learning}, which contains a convolution network and two dense layers of size (512, 512) with $tanh$ activation functions, followed by a softmax layer. For the other two tasks, we compare with a text-conditioned \textit{universal value function approximator} (UVFA)~\cite{schaul2015universal}, and the text-conditioned CNN (t-CNN) architecture of \citet{janner2017representation} which has been shown to obtain state-of-the-art performance on the PW and ISI datasets. Both models learn multi-modal representations by either concatenating text and observation vectors or using text vectors as a kernel in a convolution operation over the observation. Further details on the models are in Appendix~\ref{section:implmentation}.


\section{Results}
\label{sec:results}

\begin{figure*}[t]
\centering
\subfloat[Observational noise]{\raisebox{+2.1ex}{\includegraphics[width=2.6in]{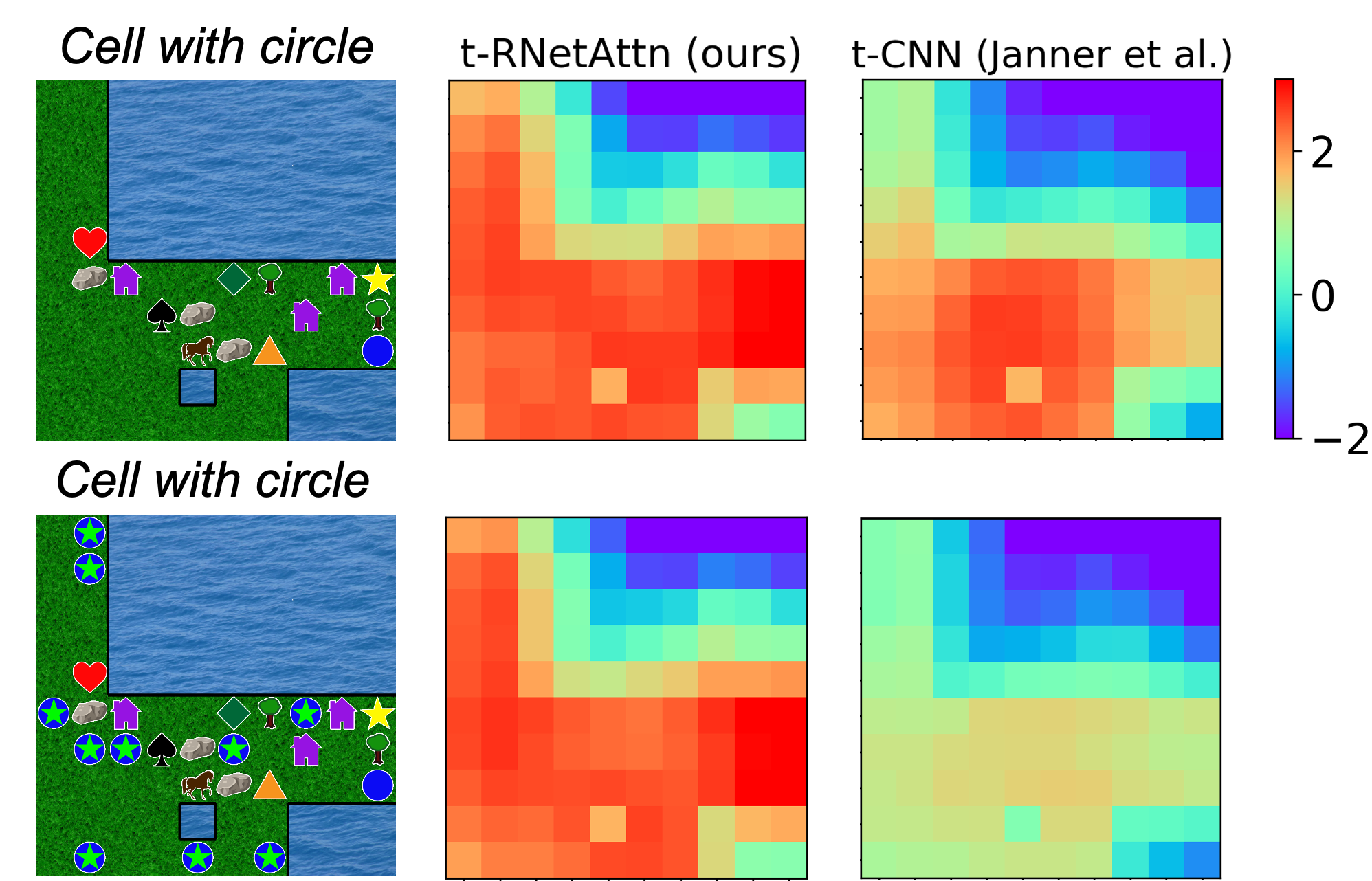}}}
\subfloat[Textual noise]{\raisebox{-0.0ex}{\includegraphics[width=2.8in]{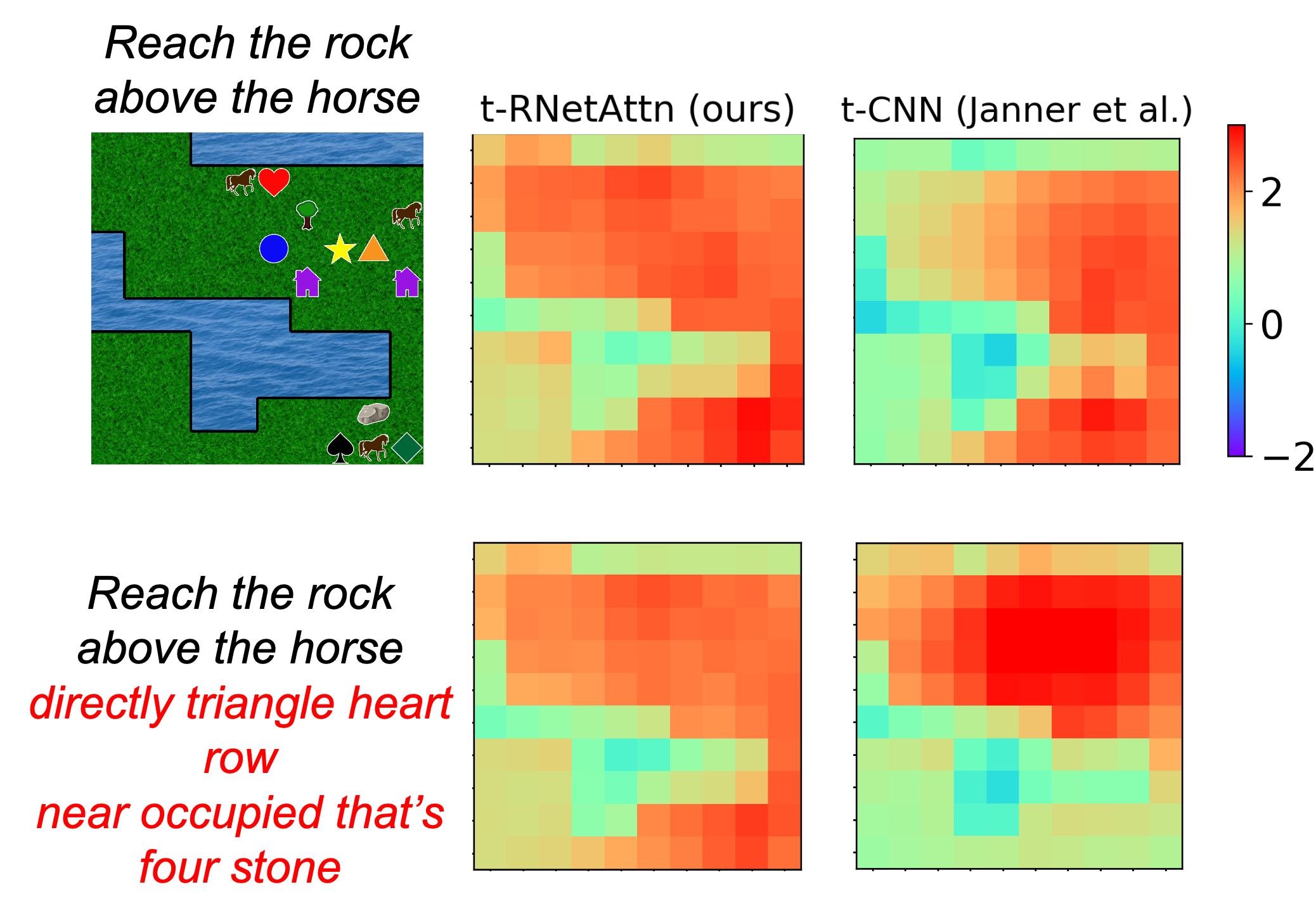}}}
\vspace{-0.1in}
\caption{Visualization of value maps from t-RNetAttn and t-CNN in PuddleWorld, without (top) and with observational and textual noise (bottom). Blue stars with circle are unseen objects which are not presented during training. t-RNetAttn is more robust and exhibits less degradation in the value map compared to t-CNN.
}
\label{fig:noiseCaseEnv}
\end{figure*}

\subsection{Overall performance}
\label{sec:Overall_performance}
Table~\ref{table:results} details the performance of our model t-RNetAttn, along with the baselines for all three tasks. Our model obtains a slightly higher accuracy of 72\% on classification compared to 71\% by t-VGG, and outperforms the baselines in MSE and policy quality (PQ) in all settings. Under MD, our model is competitive with the baselines and achieves significantly higher scores in some cases.
In the RL task, the performance gap is particularly pronounced for ISI (0.84 for t-RNetAttn vs. 0.74 for t-CNN in policy quality).  
For PuddleWorld, we observe that t-RNetAttn achieves better policy quality than t-CNN. 
This is because that RNet computes the relation score of the pair of objects individually without using a convolution kernel that takes every object into account as in t-CNN.
This allows our model to capture spatial relations at a fine-grained level. 

In the value map regression task, we observe that the proposed model achieves better performance six out of eight across all metrics.
Compared to the RL setting, we find that the regression setting achieves better performance.
This is because the RL setting requires better exploration and more interaction with the environment \ie it is harder than supervised learning of the value maps.
Overall, these observations shows that the proposed model achieves superior or competitive performance.


\subsection{Interpretability}
\label{subsection:Interpretability}
We now provide some insight into the inner workings of our model by analyzing the intermediate representations produced by both the LSTM and the relation network. Here, we focus on models trained for PuddleWorld; additional analyses for other domains are provided in the appendix.

\begin{figure}[h]
\centering
\vspace{-0.0in}
\subfloat[Words]{
\includegraphics[width=1.47in]{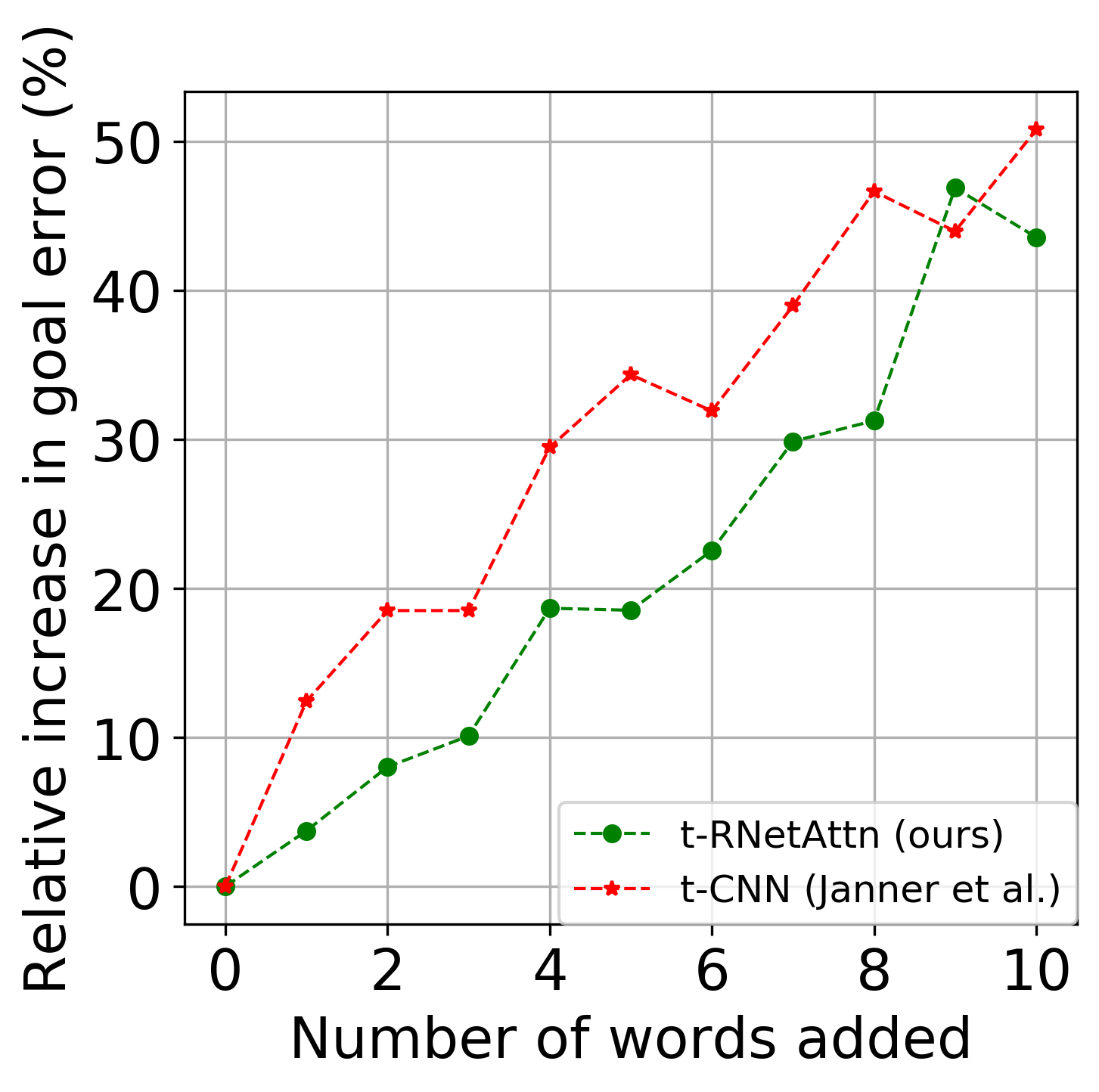}}
\subfloat[Unseen objects]{
\includegraphics[width=1.5in]{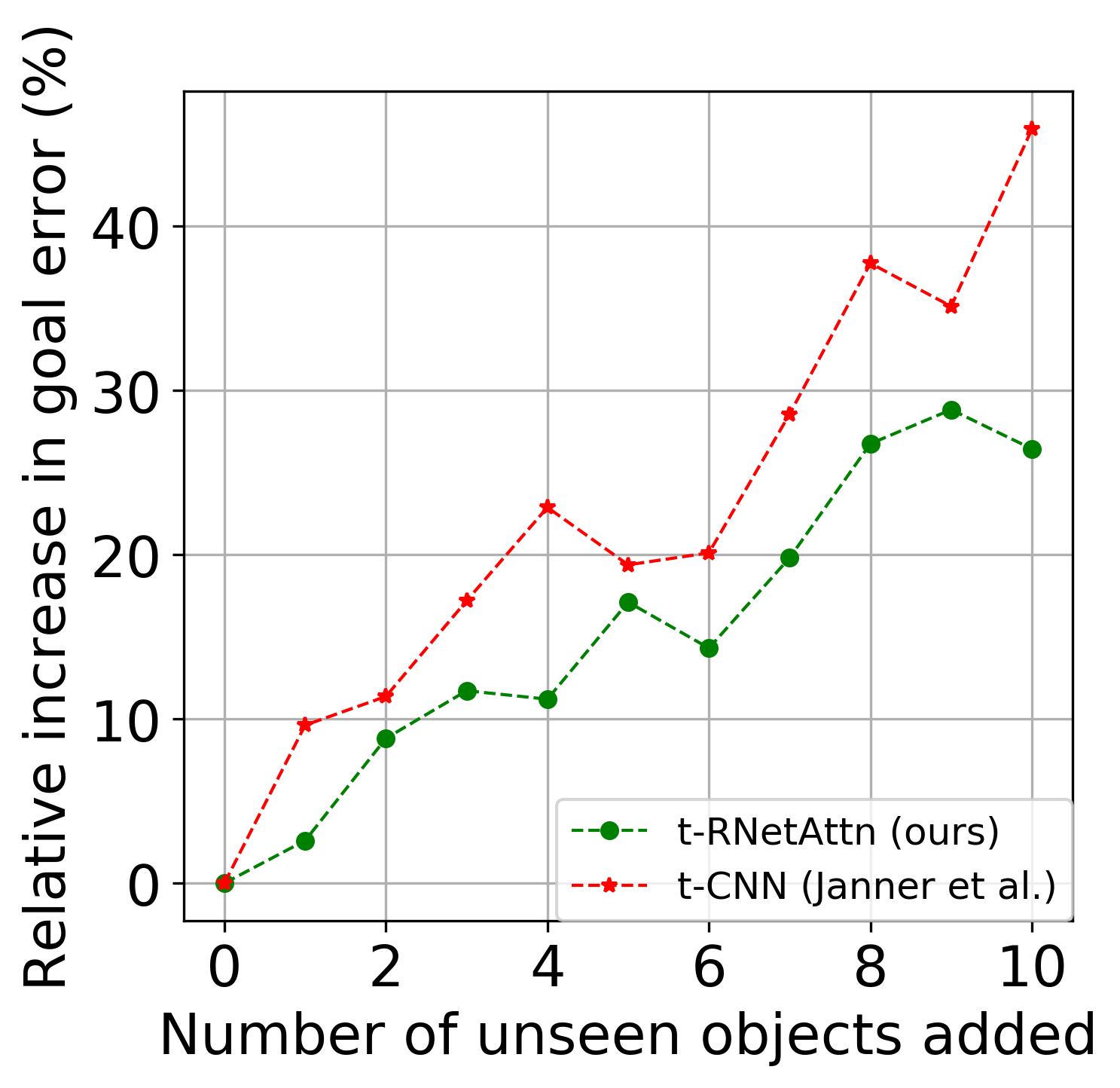}}
\vspace{-0.1in}
\caption{Relative robustness of t-RNetAttn and t-CNN under observational and textual noise in PuddleWorld, in terms of increase in goal localization error for RL goal navigation.
Our model is more robust.
}
\label{fig:relativeRobustness}
\end{figure}

\parab{Text Encoder Attention.} To understand how the text encoder handles the spatial information given by the instructions, we visualize the attention weights over the instruction text. Figure~\ref{fig:localEmbedding} shows an example of attention weights conditioned on two different locations of rocks in the environment. We observe that the model assigns higher attention weights to the correct rock instance (right), where the rock appears below a spade, thereby demonstrating correct grounding for both relevant objects in the instruction.

\parab{Relation Map.} We also visualize the value map and the relation map $|Z_{1}|$ produced by the RNet module. Figure~\ref{fig:noiseCaseEnv_inter} shows two examples from PuddleWorld (top row for both (a) and (b))). We observe that the relation network assigns sharper weights (in absolute magnitude) in $|Z_1|$ to objects mentioned in the text as well as their neighboring cells. For instance, in the first example \textit{"Reach cell one bottom and right of blue circle,"} only the circle on the map is referenced by the instruction and $|Z_1|$ shows most extreme weights for that circle.

\subsection{Robustness}
\label{subsection:robustness}
Next, we investigate the performance of our model under two kinds of noise:\vspace{0.05in} \\ 
1. \textit{Observational noise:} Here, we add (up to 10 different) unseen objects which are not presented during training to the observations at test time. Models that can ignore such objects while computing representations will be more robust to this type of noise.\vspace{0.05in} \\ 
2. \textit{Textual noise:} We also add random words, \textit{unrelated} to goal locations, into the instruction text. We randomly choose one position in the text instruction, and insert 1 to 10 words including verbs (\eg locate, reach, go), articles (\eg the, a), and irrelevant objects (\eg car, stone). We aim to test the ability of a model to ignore unhelpful words and focus on the informational content in the text.

Figure~\ref{fig:relativeRobustness} plots the relative increase in goal error (MD) for both t-CNN and t-RNetAttn as a function of the amount of observational or textual noise. 
We see that our model (green line) suffers less from both types of noise, with a drop of 30\% vs. $>45$\% for t-CNN under observational noise. 
This fact is further highlighted by Figure~\ref{fig:noiseCaseEnv} which shows the change in value maps when noise is added to both models. 
While the value maps of t-CNN change drastically, our model is less affected, especially in the prediction of the goal location (highest value). 
This observation is also strengthened by Figure \ref{fig:noiseCaseEnv_inter} which shows the change in relation map when noise is added.
We observe that in Figure \ref{fig:noiseCaseEnv_inter}(a) the relation maps are similar except that there are small values on the top-right corner of the map. 
In addition, in Figure \ref{fig:noiseCaseEnv_inter}(b) the relation maps are almost identical.
These results demonstrate that the proposed model can focus on the relevant parts of the observation map. 
In contrast, t-CNN computes a coarser global representation by taking every nearby object on the map into account. 
A small noise leads to a failure in capturing correct multi-modal representations. 
The computational costs, the failure cases, and border impact are provided in Appendix \ref{section:additionalExp}.

\section{Conclusion}
\label{sec:con}
We have presented an approach to learn robust and interpretable models for handling spatial references in text. %
We use a text-conditioned relation network to capture fine-grained spatial concepts between entity pairs, with dynamically computed weights using a cross-modal attention layer. 
Our empirical experiments over various domains demonstrate that our model matches or outperforms existing state-of-the-art systems on several metrics, e.g. achieving up to 16.7\% improvement in goal localization error. 
Further, we show that our approach is more robust to noise compared to the baselines, in terms of both unseen objects (observational noise) and randomly injected words (textual noise). 
Finally, we demonstrate that our model's intermediate representations provide a way to interpret its predictions.
Future research can explore other types of spatial relations as well as techniques to scale relation networks to larger observations spaces (\eg egocentric vision tasks) in a computationally efficient manner. 
In addition, our approach can be readily extended to more scenarios by considering longer-range cells and incorporating the object detector to extract the object in the scene.
%



\section*{Acknowledgments}
The authors would like to thank the anonymous reviewers, the area chairs, Ameet Deshpande, Michael Hu, and Princeton NLP members for their valuable feedback.

\bibliography{emnlp2020}
\bibliographystyle{acl_natbib}
\balance
\newpage
\clearpage

\appendix
{\Large\centerline{\textbf{Appendix}}}
\paragraph{Outline.}
Section \ref{section:implmentation} describes the architecture of the models, the implementation details, the instructions for executing the code, and computational resources. 
Section \ref{section:stats} details three datasets considered in the paper.
Section \ref{section:additionalExp} provides an additional analysis of interpretability and robustness in ISI and ShapeWorld. 
In addition, we discuss the potential weakness of the proposed approach.
Finally, Section \ref{section:borderImpact} discusses the border impact and the risk of deploying the proposed approach in real applications.

\section{Implementation Details}
\label{section:implmentation}

\paragraph{Architectures.}
We briefly describe the architectures of t-RNetAttn, t-CNN, and t-UVFA.

\paragraph{(1) t-RNetAttn:}
In PuddleWorld the relation network is a multilayer perceptron (MLP) with layers of $\{10, 10, 1\}$ neurons. 
We use tanh as an activation function. 
The size of each object embedding $[\phi(e)]$ is 7.
This results in a size 263 for the text vector $h$
(This is because that $(7\cdot2+1)\cdot10+10\cdot10+10\cdot1+3=263,$ where the last 3 components of $h$ are used for a gradient map).
The text encoder is an LSTM with the size 15 for input layers and the size 30 for the hidden layers, followed by a linear decoder.
In the first 260 components of $h,$ the first 150 components are reshaped into a 2-D matrix $W_1\in\mathbb{R}^{10\times15}$ for the first layer, 
the following 100 components are reshaped into a 2-D matrix $W_2\in\mathbb{R}^{10\times10}$ for the second layer, 
and the rest of the components are reshaped into a 2-D matrix $W_3\in\mathbb{R}^{1\times10}$ for the final output layer.
As a result, we compute $r_{i,j}$ by
\begin{align}
r_{i,j}&=f([\phi(s_i),\phi(s_j),l_{i,j}])
\\\nonumber
&=W_3(\sigma(W_2\sigma(W_1[\phi(s_i),\phi(s_j),l_{i,j}]))),\nonumber
\end{align}
where $\sigma(\cdot)$ is an activation function.
The remaining 3 components are used for the gradient map $Z_2.$
Finally, the relation map $Z_1$, and the gradient map $Z_2$ are concatenated, resulting in the size $(10,10,2)$ tensor, where the first two numbers are the size of the map and the last number is the size of the channel.
We then use a convolution operation with the kernel size 3 and a relu activation function to get the final value maps.

In ISI we use the same architectures in PuddleWorld except for $[\phi(e)]$ being 13. 
This results in a size 383 for text vector $h.$
(This is because that $(13\cdot2+1)\cdot10+10\cdot10+10\cdot1+3=383,$ where the last 3 components of $h$ are used for a gradient map.)

In ShapeWorld we use the same architectures in PuddleWorld except for replacing the final convolution layer with a dense layer followed by a softmax function to predict labels.

\paragraph{(2) t-CNN:} 
In PuddleWorld we use a convolution filter with a size $(3,3)$. 
(We also increase the size of filters to match the number of parameters in the proposed model. We find that the baseline performance drops by 5\% since there are too many entities on a single filter.)
This results in a size 66 for $h$
(This is because that $3\cdot3\cdot7+3=66,$ where the last 3 components of $h$ are used for a gradient map).
We use the first 63 components of $h$ to be a convolution kernel of size $(3,3,7)$ on an environment map, where the value $7$ is the size of the object embedding. 
This results in a text-conditioned map $Z_1$ with a size $(10,10,1)$
The remaining components are used to construct a gradient map.
Finally, the relation map $Z_1$, and the gradient map $Z_2$ are concatenated, resulting in the size $(10,10,2)$ tensor.
We then use a convolution operation with the kernel size 3 and a relu activation function to get the final value maps.
    
In ISI we use the same architectures in PuddleWorld. We use a convolution filter with a size 3.  This creates a size 120 for $h$ (This is because that $3\cdot3\cdot13+3=120$).

\begin{figure*}[t]
\centering
\vspace{-0.0in}
\subfloat[Local instruction]{\includegraphics[width=2.0in]{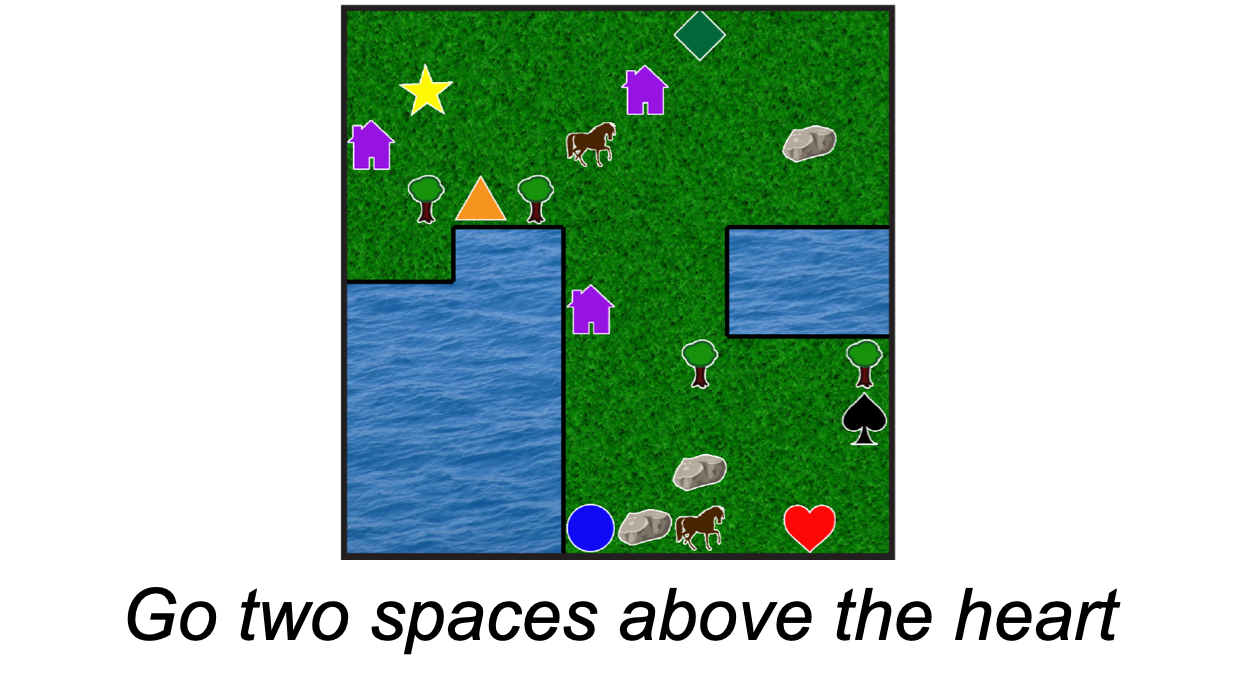}}
\subfloat[Global instruction]{\includegraphics[width=2.0in]{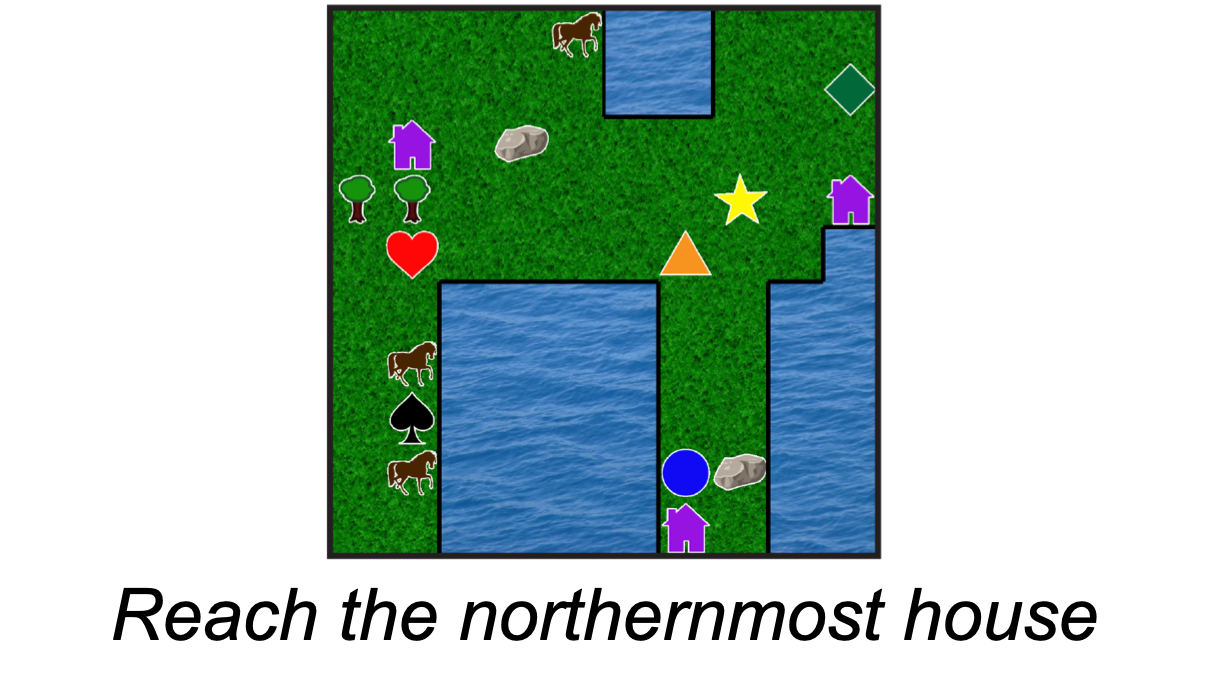}}
\vspace{-0.1in}
\caption{Examples of the local and the global instructions in PuddleWorld. 
A local instruction specifies the goal by using nearby objects.
A global instruction specifies the goal by using a global viewpoint.
}
\vspace{-0.1in}
\label{fig:pw_local_global}
\end{figure*}

\begin{figure*}[t]
\centering
\vspace{-0.0in}
\subfloat[]{\includegraphics[width=2.25in]{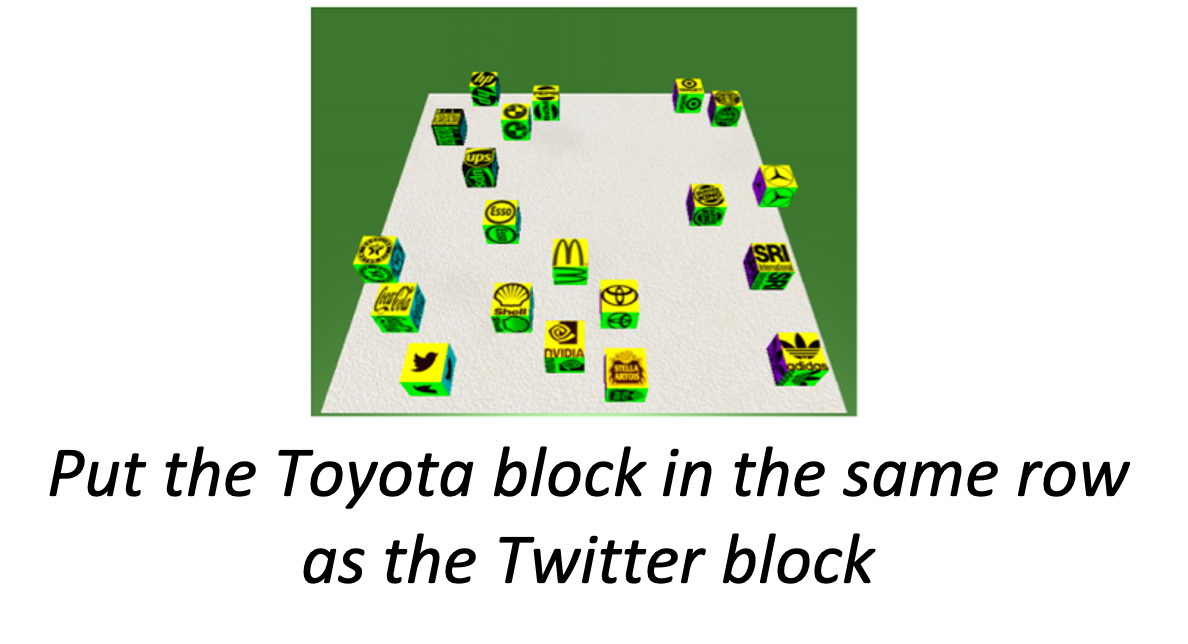}}
\subfloat[]{\includegraphics[width=2.55in]{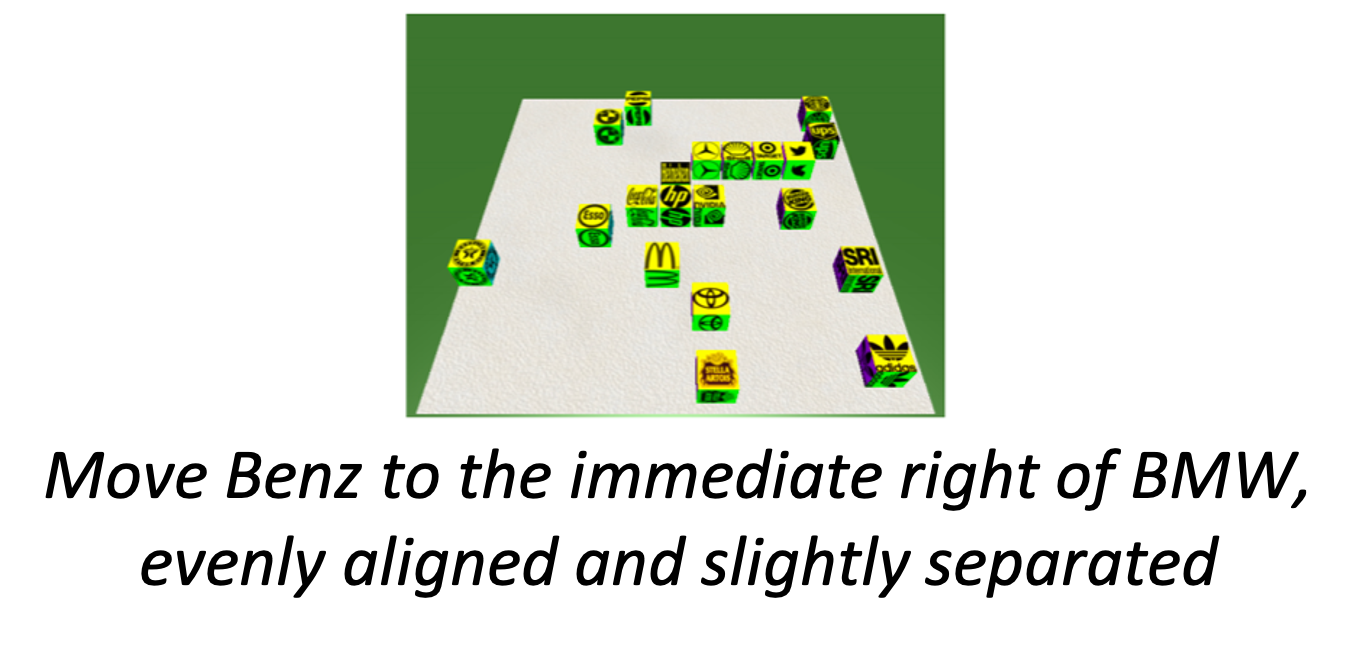}}
\vspace{-0.1in}
\caption{Examples in ISI. The goal is to place the block in the right position specified by the instructions.
}
\vspace{0.0in}
\label{fig:isi_demo}
\end{figure*}

\paragraph{(3) t-UVFA:} For all three datasets, the size of $h$ is 7, followed by concatenating $h$ and state representations for the map to obtain a multi-modal representation. This representation is then fed into a deconvolution layer used to decode and reconstruct value maps.

For all the models and datasets, we use the batch size 512. 
The step size is $10^{-3}$.
We choose Adam algorithm to train the models.

\paragraph{Hyperparameter Search.}

We conduct a grid search on the embedding size of the object $\{4,5,6,7,8,9,10\}$, the step size $\{10^{-1},10^{-2},10^{-3},10^{-4}\}$, the batch size $\{100,500,1000\}.$ 
We test the performance (\ie the policy quality) of each combination of the hyperparameter on the validation set separated from the training dataset. 

\paragraph{Instructions for Reproducibility.} 
To reproduce the results,  
first install the libraries for python3 such as numpy, scipy and PyTorch. 
Then download the package from \url{https://github.com/anonymous}.
Put the code on the designated folder.
Finally, go to the example folder and execute the code using python command.

\paragraph{Computational Resources.} We conduct the experiments on a machine with an Intel Core i7 CPU and no GPU is used. 
We find that in general it takes about 6 hours to complete the experiments across all three models. 
This shows that the proposed model does not have a substantial computation cost compared to the other baselines.
All experiments are performed using PyTorch\footnote{\url{https://pytorch.org}}.
%

\section{Datasets}
\label{section:stats}

\paragraph{PuddleWorld.} 
A 10 $\cdot$ 10 grid represents the states. The cell is placed with either a water or a grass.
In addition, a grass may be placed with six unique objects (triangle, star, diamond, circle, heart, and spade) appearing once per map or the non-unique objects (rock, tree, horse, and house).
Two types of instructions are provided: a local instruction that describes the goal location with the nearby objects, \eg ``\textit{two cells to the left of the triangle},'' and a global instruction that specifies the goal location with the global viewpoint, \eg ``\textit{the westernmost rock}.''
Figure \ref{fig:pw_local_global} shows the examples of the local and global instructions.
We give a reward of $+3$ when the agent reaches the final goal location.
The reward design is to encourage the agent to produce the accurate value maps.
Please refer to \citet{janner2017representation} for more discussion on data collection process. 

\begin{figure*}[t]
\centering
\vspace{-0.0in}
\subfloat[Observational noise]{
\includegraphics[width=3.1in]{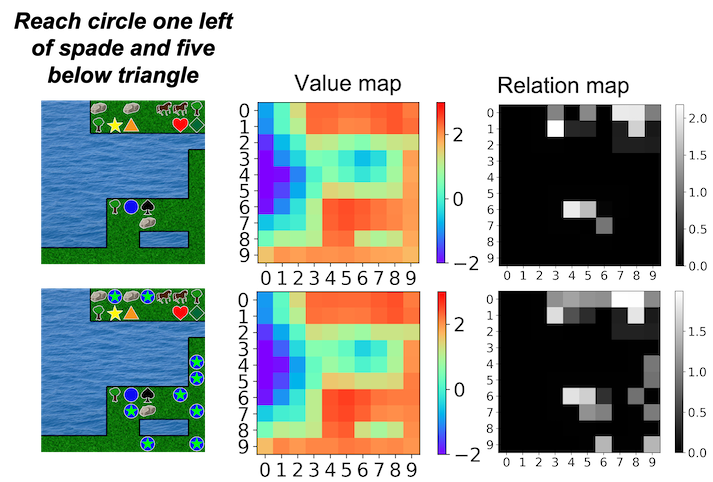}}
\subfloat[Observational noise]{
\includegraphics[width=3.1in]{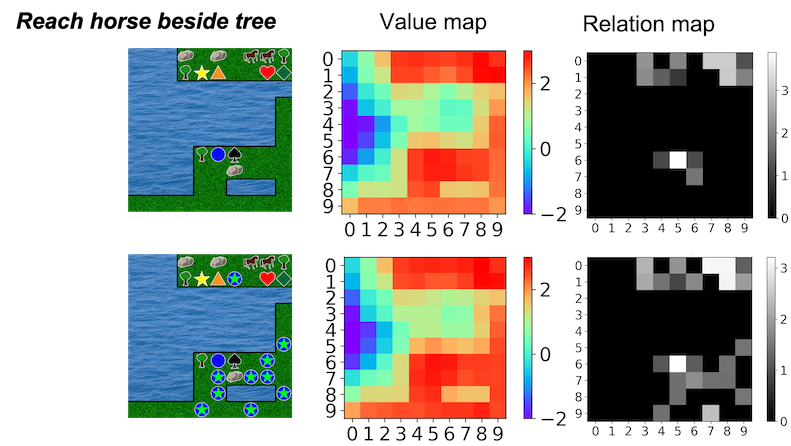}}

\subfloat[Textual noise]{\includegraphics[width=3.1in]{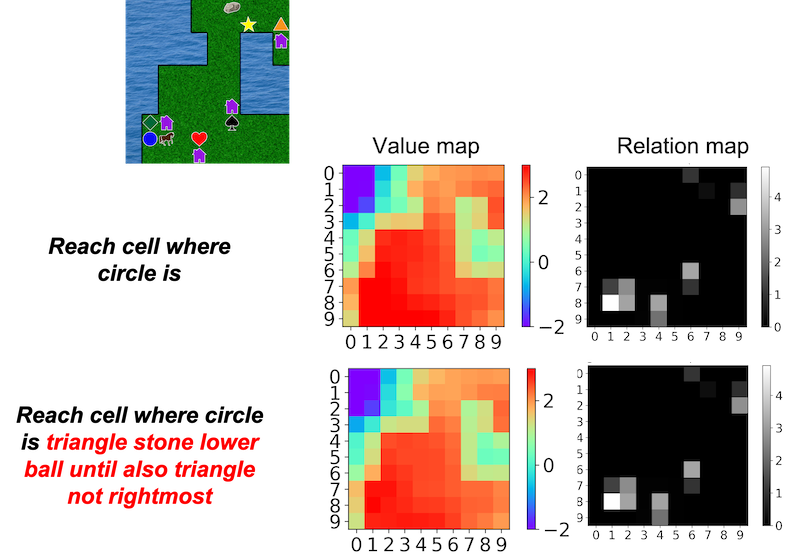}}
\subfloat[Textual noise]{\includegraphics[width=3.1in]{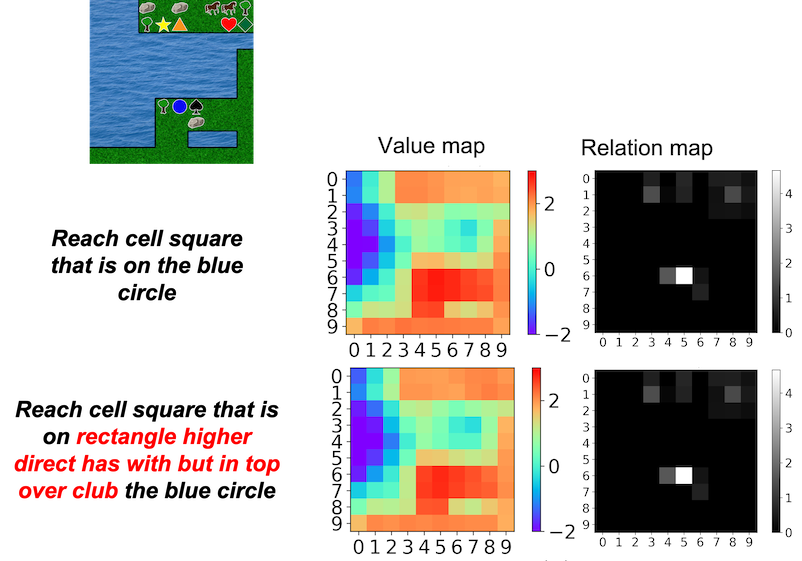}}
\vspace{-0.1in}
\caption{Visualization of value maps and relation maps after taking absolute values $|Z_1|$ from t-RNetAttn, without (top) and with observation and textual noise (bottom) in the PuddleWorld environment. Blue stars are unseen objects. Our approach produces sharper (magnitude-wise) $|Z_1|$ values for goal location and referent objects, and is almost undisturbed by noise.}
\label{fig:noiseCaseEnv_appendix_}
\end{figure*}

\paragraph{ISI.}
The environment contains up to 20 blocks marked with logos (\eg Toyota, BMW) or digits. 
Each instruction specifies the goal location of the object, \eg ``\textit{Move Toyota to the immediate right of SRI, evenly aligned and slightly separated}.''
Figure \ref{fig:isi_demo} shows the examples.
Please refer to \citet{bisk2018learning} for more discussion on data collection process. 

\paragraph{ShapeWorld.}
Each scene contains 4 or 5 non-overlapping objects. 
Unlike the object in the Puddle and ISI that has a unique identifier, the object in the ShapeWorld is a pixel image.
This is to demonstrate that the proposed approach can operate on the raw images.
The instruction describes the spatial relationships between pairs of objects specified by shape, color, or both, \eg ``\textit{a red ellipse is to the right of an ellipse}.''
There are 8 colors and 8 shapes in total. 
Unlike the previous two tasks that predict a target location, the task in the ShapeWorld is to classifier whether the instruction matches the scene.
Figure \ref{fig:example} (bottom) shows the examples.
Please refer to \citet{andreas2017learning} for more discussion on data collection process.

\paragraph{Dataset Splitting.} We follow the same splitting scheme as in \citet{janner2017representation,bisk2018learning,andreas2017learning}.
We show dataset statistics in table \ref{table:dataset}.

\begin{table}[h]
\centering
\vspace{0.0in}
\centering
\scalebox{0.8}{
\begin{tabular}{ccc}
\toprule
                          Dataset        & Train   & Test   \\  \hline
     \multirow{1}{*}{PW local}   & 1566   & 399   \\  
     \multirow{1}{*}{PW global} &  1071  & 272   \\
     \multirow{1}{*}{ISI}             & 11871 & 3177 \\ 
     \multirow{1}{*}{SW}      & 9000  &  500\\ 
   \bottomrule
\end{tabular}}
\vspace{-0.1in}
\caption{Statistics of PuddleWorld (PW), ISI language grounding (ISI), and ShapeWorld (SW).}
\label{table:dataset}
\end{table}

\section{Additional Results}
\label{section:additionalExp}

\begin{figure*}[t]
\centering
\vspace{-0.0in}
\subfloat[]{
\includegraphics[width=0.48\linewidth]{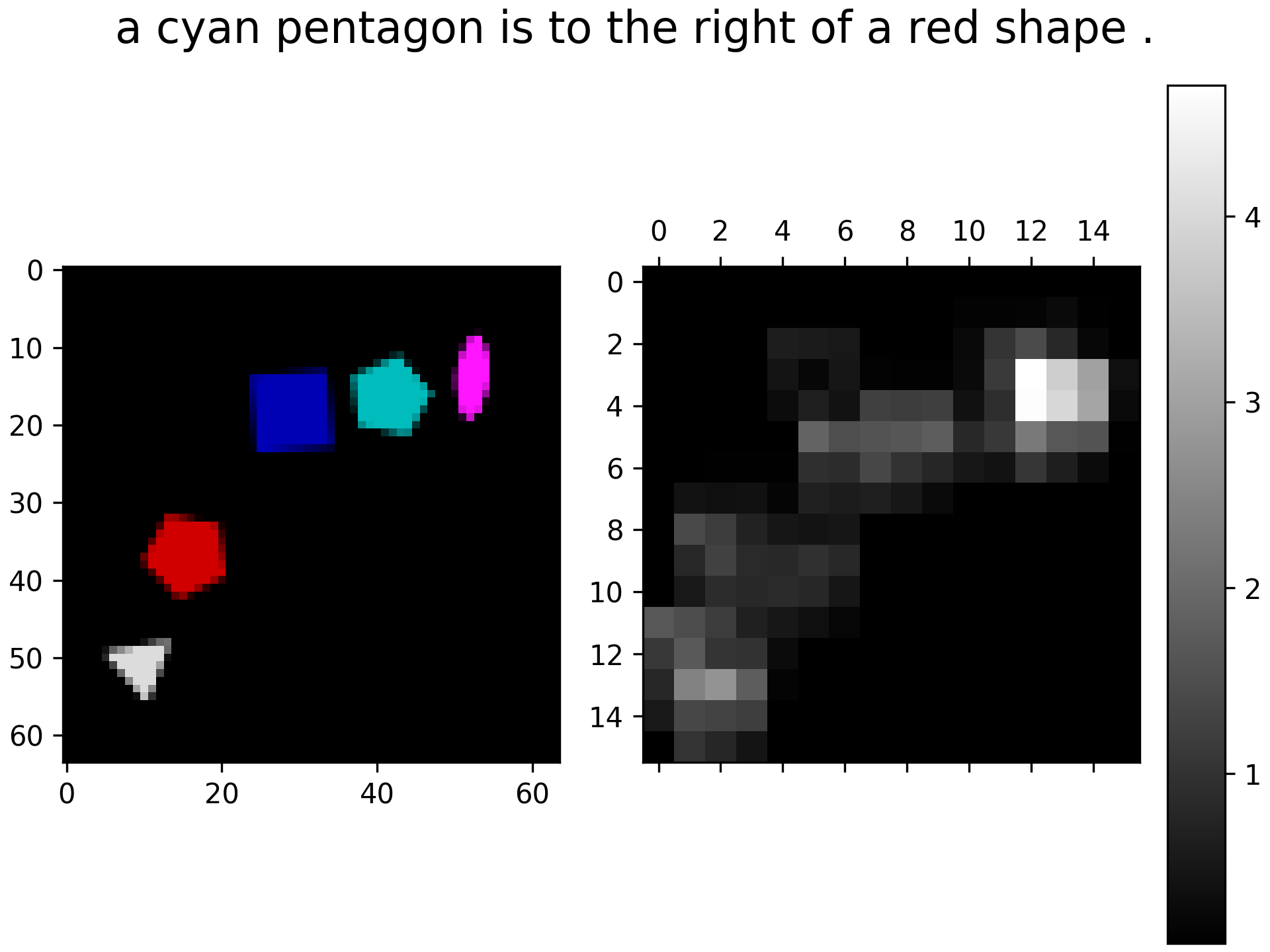}}
\subfloat[]{
\includegraphics[width=0.48\linewidth]{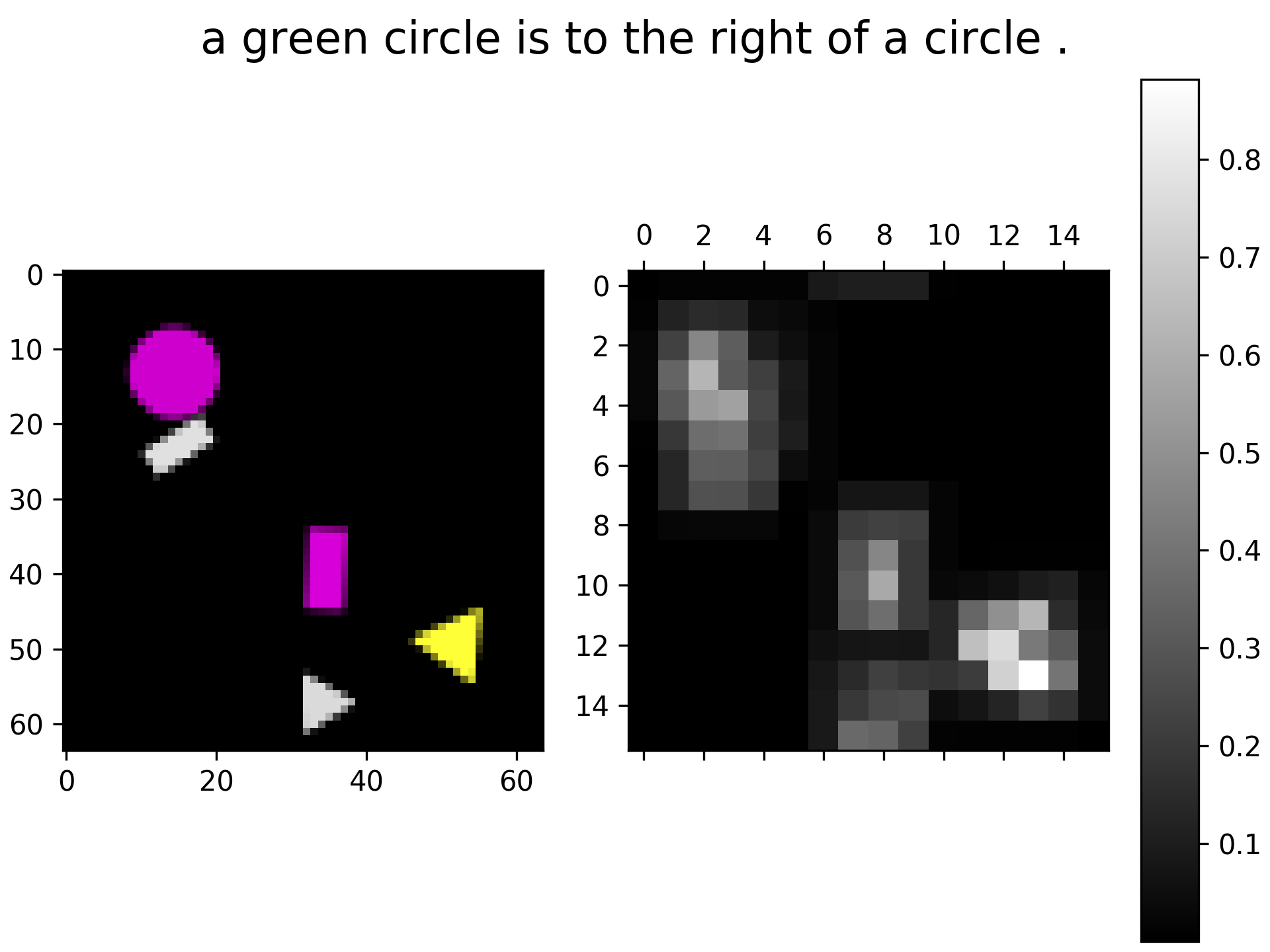}}
\vspace{-0.1in}
\caption{
Visualization of value maps and relation maps after taking absolute values $|Z_1|$ from t-RNetAttn in the ShapeWorld environment. 
Our approach produces sharp (magnitude-wise) $|Z_1|$ values for goal location and referent objects.
Note that the size of the environment map is different from the size of the relation map since we reduce the size of the environment map by using a convolution operation, reducing the computational cost.
}
\label{fig:noiseCaseEnv_appendix_SW}
\end{figure*}

\begin{figure*}[t]
    \centering
    \begin{minipage}{0.45\textwidth}
        \centering
\subfloat[Observational noise]{ 
\includegraphics[width=1.4in]{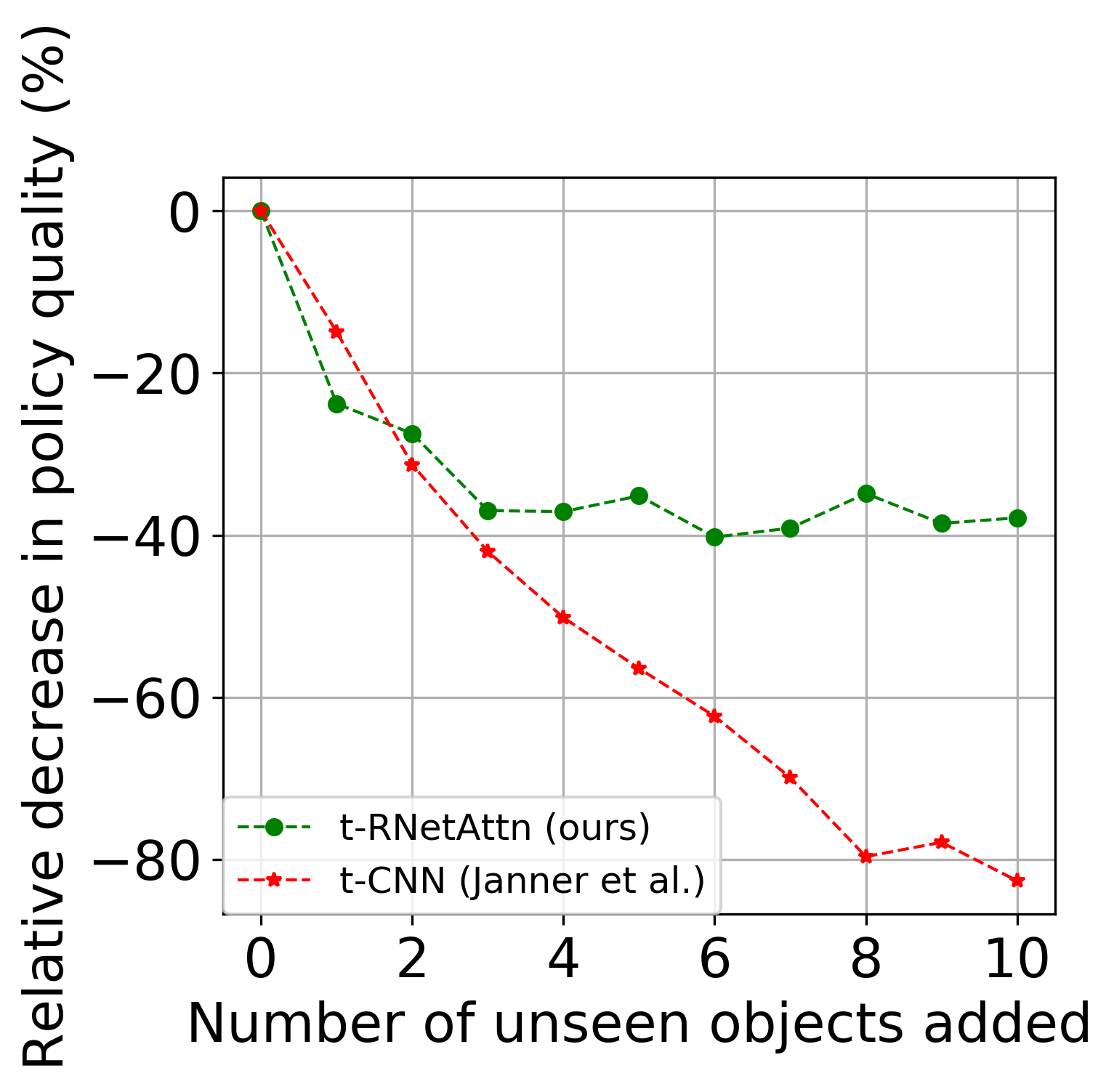}}
\subfloat[Textual noise]{
\includegraphics[width=1.4in]{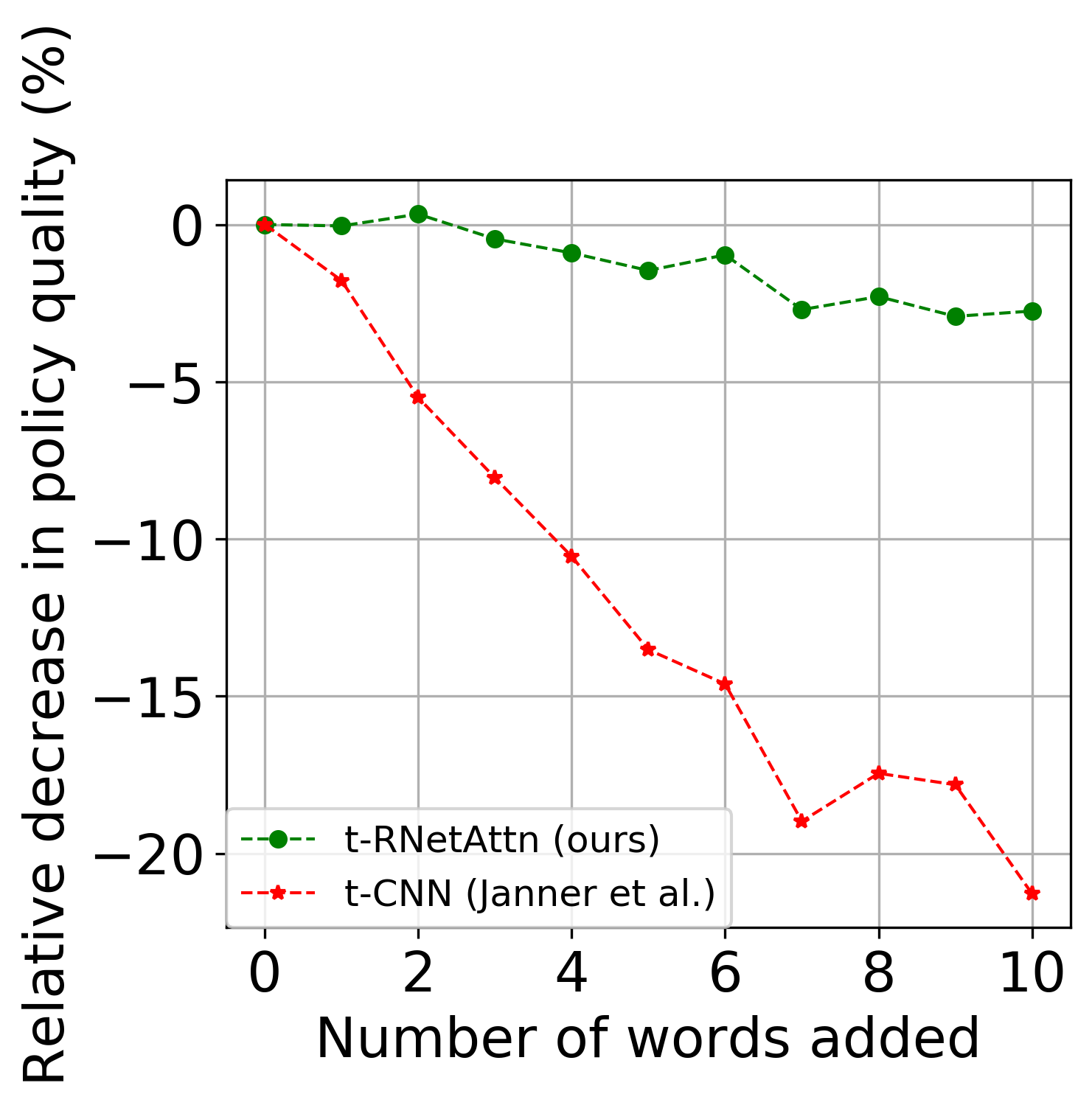}} 
        \caption{Relative robustness of t-RNetAttn and t-CNN under observational and textual noise in ISI, in terms of decrease in policy quality for goal navigation with RL.}
        \label{fig:relativeRobustness_ISI}
    \end{minipage}\hfill
    \begin{minipage}{0.45\textwidth}
        \centering
\subfloat[Observational noise]{
\includegraphics[width=1.4in]{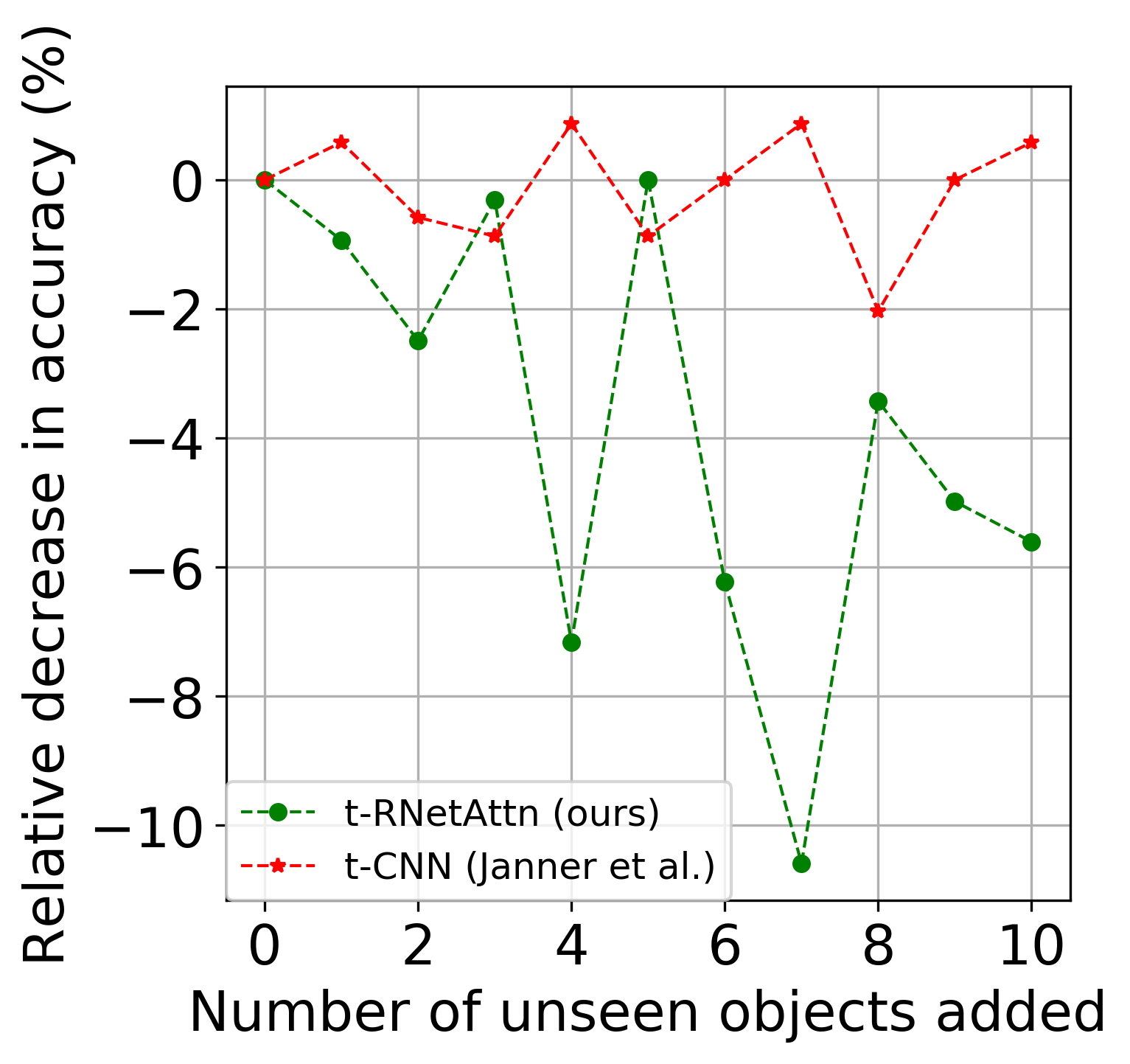}}
\subfloat[Textual noise]{
\includegraphics[width=1.4in]{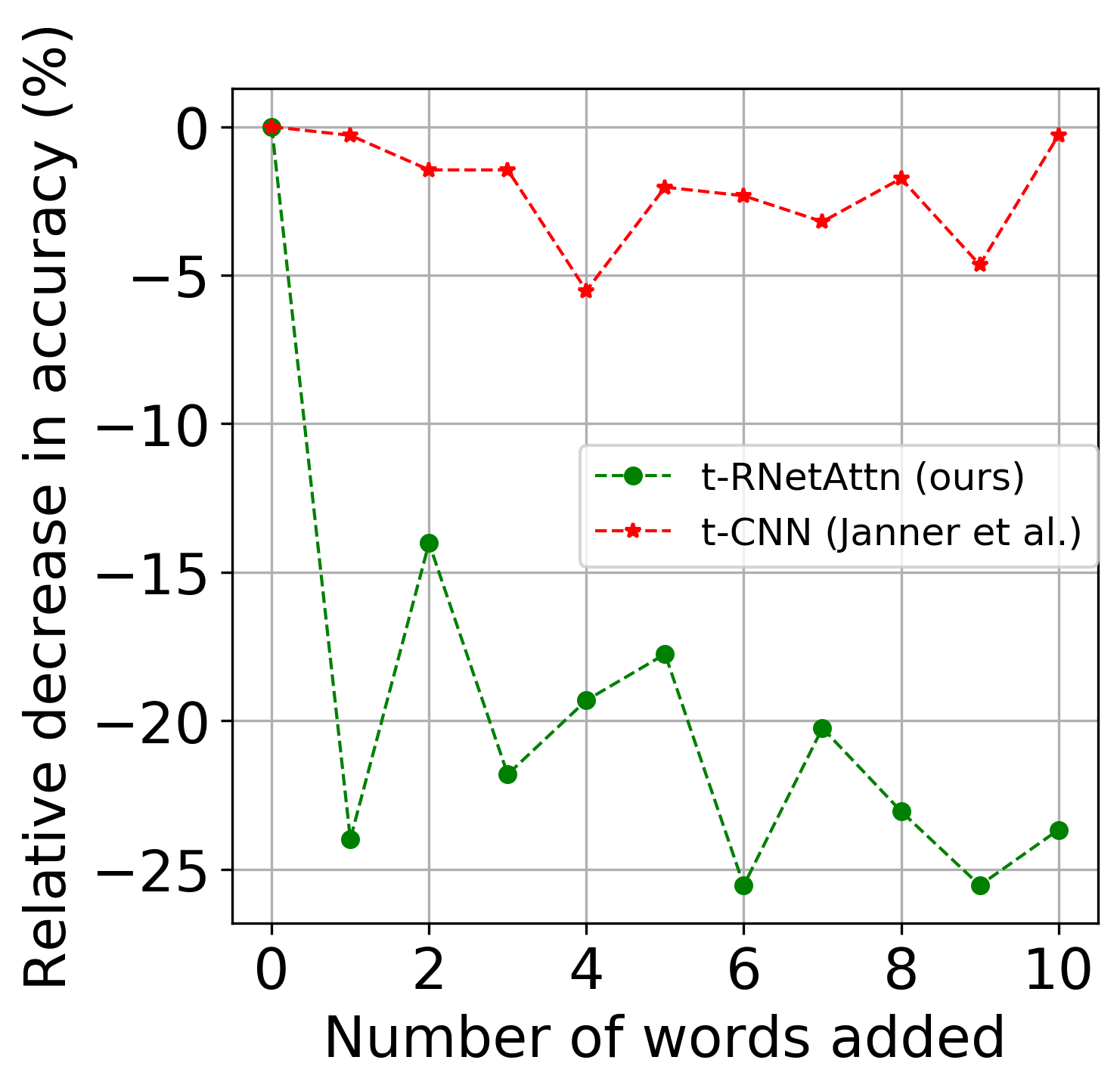}} 
        \caption{Relative robustness of t-RNetAttn and t-CNN under observational and textual noise in ShapeWorld, in terms of decrease in prediction accuracy.}
        \label{fig:relativeRobustness_SW}
    \end{minipage}
\end{figure*}

\subsection{Interpretability}
\label{appendix:Interpretability}
\paragraph{PuddleWorld.}
To show the proposed model can increase the interpretability, we provide additional visualization examples of relation map in PuddleWorld as shown in Figure \ref{fig:noiseCaseEnv_appendix_}. 
We observe that the proposed model assigns a larger magnitude of the weights to the objects mentioned in the text in the relation map.
For example, in Figure \ref{fig:noiseCaseEnv_appendix_}(d) we observe that the model successfully attends to ``\textit{circle}'', which is specified by the instruction.
In addition, under the observational and textual noise the value maps are almost undisturbed by noise.
For example, in Figure \ref{fig:noiseCaseEnv_appendix_}(a) and (b) we observe that the relation maps are almost unchanged after adding observational noise except for relatively small values on the unseen object. 
On the other hand, in Figure \ref{fig:noiseCaseEnv_appendix_}(c) and (d) we observe that the relation maps are almost identical after adding textual noise. 
These observations imply that explicitly computing the relation score of each entity pairs allows the model to attend to objects that are most relevant to the instruction.     
The results here are also consistent with the results in Figure \ref{fig:noiseCaseEnv_inter}. 

\paragraph{ShapeWorld.}
To show the proposed model can increase the interpretability, we provide visualization examples of the relation map in ShapeWorld as shown in Figure \ref{fig:noiseCaseEnv_appendix_SW}. 
We observe that t-RNetAttn also assigns a larger magnitude of the weights to the objects mentioned in the text.
For example, in Figure \ref{fig:noiseCaseEnv_appendix_SW}(a) we observe that there is a larger magnitude of the weights on the top-right corner of the map.
This implies that the proposed model successfully attends to the object that is specified by the instruction (``\textit{a cyan pentagon}'').
This observation shows that the proposed model still works well when the raw images are presented.  
%
%

\subsection{Robustness}
\label{appendix_Robustness}

%
%

\begin{table*}[t]
\centering
\vspace{0.0in}
\centering
\subfloat[Classification]{
\scalebox{0.75}{
\begin{tabular}{l*{1}{c}r}
 &\multicolumn{1}{c}{SW}\\ \cmidrule(lr){2-2} & ACC $\uparrow$\\ \hline
t-RNet (ours)  & \bf{0.73}\\ 
t-RNetAttn (ours)  & {0.72}\\  
\bottomrule
\end{tabular}
}}
\subfloat[Value map regression]{
\scalebox{0.75}{
\begin{tabular}{l*{8}{c}r}
 &\multicolumn{3}{c}{PW local}
                                    &\multicolumn{3}{c}{PW global}  
                                    &\multicolumn{2}{c}{ISI}  \\ \cmidrule(lr){2-4} \cmidrule(lr){5-7}
                                    \cmidrule(lr){8-9}
                       & MSE  $\downarrow$ & PQ $\uparrow$ & MD $\downarrow$  & MSE  $\downarrow$ & PQ $\uparrow$ & MD $\downarrow$ & MSE  $\downarrow$ & MD $\downarrow$ \\ \hline
                                                            
t-RNet (ours)                             & \bf{0.19} & \bf{0.94} & \bf{1.95}  &\bf{0.31}& \bf{0.91} & \bf{3.13} & {0.16}  & 3.66 \\ 
t-RNetAttn (ours)                             & {0.22} & 0.92 & \bf{1.95}  &{0.40}& \bf{0.91} & 3.82 & \bf{0.15}  & 3.43 \\

\bottomrule
\end{tabular}
}

}

\subfloat[Goal navigation with RL]{
\scalebox{0.75}{
\begin{tabular}{l*{6}{c}r} 
&\multicolumn{2}{c}{PW local} &\multicolumn{2}{c}{PW global} &\multicolumn{2}{c}{ISI }   
\\ \cmidrule(lr){2-3} \cmidrule(lr){4-5}\cmidrule(lr){6-7}
& PQ $\uparrow$ & MD $\downarrow$  & PQ $\uparrow$ & MD $\downarrow$ & PQ $\uparrow$ & MD $\downarrow$ \\ \hline
t-RNet (ours)                             & \bf{0.92} & 2.41  & \bf{0.93} & \bf{3.56} & \bf{0.88} & \bf{3.22} \\
t-RNetAttn (ours)                             & {0.91} & 2.10  & \bf{0.93} & 4.23 & {0.84} & {3.79} \\

\bottomrule
\end{tabular}}
}

\vspace{-0.1in}
\caption{Performance on the test set with the three metrics (PQ: policy quality, MD: Manhattan distance, MSE: mean squared error) in PuddleWorld, ISI and ShapeWorld under both supervised and RL. The symbols $\uparrow$ and $\downarrow$ signify larger numbers are better and smaller numbers are better, respectively. The best values are bold. Note that in the case of supervised learning, we do not report the value of t-VGG since it is designed to solve the task in ShapeWorld (SW). The MSE of t-UVFA in the ISI is not reported since it was not reported in \citet{janner2017representation}.}
\label{table:results_appendix_with_without_Attention}
\end{table*}

\paragraph{ISI.}
To show that the proposed model is more robust to the noise, Figure~\ref{fig:relativeRobustness_ISI} plots the relative decrease in policy quality for both t-CNN and t-RNetAttn as a function of the amount of observational or textual noise in ISI. We can see that our model (green line) suffers less from both types of noise (a drop of 40\% vs. $>$80\% for t-CNN on the observational noise with 10 unseen objects and a drop of 2.5\% vs. $>$20\% for t-CNN on the textual noise with 10 random words). 
This implies that the proposed approach is robust to the noise because of the relation network.

\paragraph{ShapeWorld.}
To show that the proposed model is more robust to the noise, Figure \ref{fig:relativeRobustness_SW} plots the relative decrease in accuracy for both t-CNN and t-RNetAttn as a function of the amount of observational or textual noise in ShapeWorld. 
For the observational noise, instead of adding the unseen objects, we add noise patches in the input images. The element of each patch is sampling from Gaussian distribution with the mean being zero and variance being one. 
For the textual noise, we use the same procedure as the one in the PuddleWorld and ISI.
We can see that our model (green line) suffers more from both types of noise. 
One possible reason for this is that in order to reduce the dimension of the observation map, we first perform a convolution on the observation map.
The resulting embeddings with a smaller width and height are the inputs to the relation module.
Unlike directly using the embedding from the original observation map, this dimension-reduction approach creates a coarse representation of the observation map.
This makes a relation module vulnerable to observational and textual noise. 
In contrast, t-CNN directly operates on the observation map.
This makes t-CNN less vulnerable to the noise.
One solution to increase the robustness of t-RNetAttn in ShapeWorld is to use an object detector to segment the objects from the map. 
This would allow t-RNetAttn directly to use the object information rather than the embeddings from the pixel.
We leave the improvement of this as a future research direction.

\subsection{Ablation Study}
\label{appendix:ablation}
One question of the proposed model is that whether the improvement in performance is due to RNet or the attention.
To this end, we remove the attention mechanism and simply take an average over all LSTM outputs as $h.$
Figure \ref{table:results_appendix_with_without_Attention} shows the result.
We include the numbers reported in Table \ref{table:results} for clarity. 
%
%
%

\subsection{Limitations of the Proposed Model}
\label{appendx:weakness}
The robustness experiment in ShapeWorld in Section \ref{appendix_Robustness} shows that the proposed model is vulnerable when the embedding comes from raw pixels instead of the entity itself.
Another limitation is that the proposed model only computes representations in a 1-square neighborhood around each cell. 
This may be problematic when we ask it to resolve ``\textit{reach the cell \textbf{three above} the easternmost star}'', where the goal is three blocks away from the star.
One possible solution to this is that we can have another relation network that considers the entities in a 3-square neighborhood around each cell. 
This creates another relation map $Z_3$ (similar to $Z_1$) that captures the long dependency.
Then we concatenate $Z_3$ with $Z_1$ and $Z_2$ and feed this tensor ($[Z_1;Z_2;Z_3]\in\mathbb{R}^{m\times n\times2}$) into a convolutional layer to predict the value function.
We leave the improvement of this as a future research direction.

\section{Border Impact}
\label{section:borderImpact}
The proposed approach could be applied in many fields that are required to learn multi-modal representations while providing transparency of the model.
For example, in the personalized robotic assistants setting where the agent aims to complete tasks specified by the instruction,
the transparency in the model is critical to build trust between humans and AI and mitigate safety risks in making decisions. 
In addition, in the visual question answering \cite{antol2015vqa} where it is required to have a visual understanding of the scene to answer many questions,
the proposed model could enhance the performance while unveiling the inner work of the model.
Moreover, in the Simultaneous Localization
and Mapping (SLAM) \cite{durrant2006simultaneous} system where the agent aims to construct the map of the environment and locate itself, 
the proposed model could fuse multi-sensor input data such as laser and ultrasonic sensors to learn a representation for generating the map.

%
%
%
%

%
%
%
%


\end{document}